\pdfoutput=1

\documentclass[11pt]{article}

\usepackage[final]{acl}

\usepackage{times}
\usepackage{latexsym}

\usepackage[T1]{fontenc}
\usepackage[utf8]{inputenc}

\usepackage{microtype}

\usepackage{inconsolata}

\usepackage{graphicx}

\usepackage{xcolor}
\usepackage[table]{xcolor}

\usepackage{multirow, makecell}
\usepackage{booktabs}
\usepackage{svg}
\usepackage{amsmath}
\DeclareMathOperator*{\argmin}{arg\,min}

\usepackage{xspace}
\usepackage{multirow}
\usepackage[T1]{fontenc}
\usepackage{tcolorbox}
\tcbuselibrary{skins,breakable,listings,hooks} % no minted here; pure verbatim
\usepackage{listings}
\usepackage{tabularx}
\usepackage{graphicx}
\usepackage{subcaption}
\usepackage{tikz}
\usetikzlibrary{arrows.meta, positioning}
\lstdefinestyle{promptstyle}{
  basicstyle=\ttfamily\small,
  breaklines=true,
  breakatwhitespace=false,
  columns=fullflexible,
  keepspaces=true,
  showstringspaces=false,
  language={}, % important: no language => no % comments, no special syntax
  escapeinside={(*@}{@*)}
}
\definecolor{tealblue}{rgb}{0.0, 0.5, 0.7}
\usepackage{pifont}

\usepackage{algorithm}
\usepackage{algorithmic}
\usepackage[para]{footmisc}
\newtcblisting[auto counter, number within=section]{promptbox}[2][]{%
  enhanced,
  breakable,
  listing only,
  listing engine=listings,
  listing options={style=promptstyle},
  colback=gray!3,
  colframe=black!35,
  boxrule=0.3pt,
  sharp corners,
  left=1em, right=1em, top=0.6em, bottom=0.6em,
  title={Full Prompt~\thetcbcounter: #2},
  fonttitle=\bfseries,
  coltitle=black,
  attach boxed title to top left={yshift=-2mm, xshift=1mm},
  boxed title style={colback=white, sharp corners, boxrule=0pt},
  borderline west={2pt}{0pt}{blue!65!black},
  #1
}
\newtcblisting[auto counter, number within=section]{iclbox}[2][]{%
  enhanced,
  breakable,
  listing only,
  listing engine=listings,
  listing options={style=promptstyle},
  colback=green!2,
  colframe=green!40!black,
  boxrule=0.3pt,
  sharp corners,
  left=1em, right=1em, top=0.6em, bottom=0.6em,
  title={ICL Prompt~\thetcbcounter: #2},
  fonttitle=\bfseries,
  coltitle=black,
  attach boxed title to top left={yshift=-2mm, xshift=1mm},
  boxed title style={colback=white, sharp corners, boxrule=0pt},
  borderline west={2pt}{0pt}{green!60!black},
  #1
}

\newcommand{\0}{\phantom{0}}
\newcommand{\ours}{DTDR\xspace}  % our method name
\newcommand{\ourscl}{DTDR-C\xspace}  % our method name cluster
\newcommand{\oursnn}{DTDR-L\xspace}  % our method name NN (linear)
\newcommand{\bfu}[1]{\textbf{\underline{#1}}}
\newcommand{\spm}{$\scriptstyle{\pm}$}

\definecolor{verylightgray}{gray}{0.9}
\definecolor{darkgreen}{rgb}{0,0.5,0}

\newcommand{\funcF}{\mathcal{F}}
\newcommand{\cmark}{\textcolor{green}{\ding{51}}}  % checkmark
\newcommand{\xmark}{\textcolor{red}{\ding{55}}}    % cross
\usepackage{enumitem}
\title{Dynamic Tool Dependency Retrieval \\ for Lightweight Function Calling}

\author{
 \textbf{Bhrij Patel\textsuperscript{*1,2}},
 \textbf{Davide Belli\textsuperscript{*1}},
 \textbf{Amir Jalalirad\textsuperscript{1}},\\
 \textbf{Maximilian Arnold\textsuperscript{1}},
 \textbf{Aleksandr Ermolov\textsuperscript{1}},
 \textbf{Bence Major\textsuperscript{1}}
\\
\\
 \textsuperscript{1}Qualcomm AI Research \\
 \textsuperscript{2}University of Maryland, College Park, USA
\\
 \small{
   \textbf{Correspondence:} \href{mailto:bbp13@umd.edu}{bbp13@umd.edu}, \{\href{mailto:dbelli@qti.qualcomm.com}{dbelli}, \href{mailto:ajalalir@qti.qualcomm.com}{ajalalir}, \href{mailto:marnold@qti.qualcomm.com}{marnold}, \href{mailto:aermolov@qti.qualcomm.com}{aermolov}, \href{mailto:bence@qti.qualcomm.com}{bence}\}@qti.qualcomm.com
 }
}

\begin{document}
\maketitle
\begin{abstract}
Function calling agents powered by Large Language Models (LLMs) select external tools to automate complex tasks. On-device agents typically use a retrieval module to select relevant tools, improving performance and reducing context length. However, existing retrieval methods rely on static and limited inputs, failing to capture multi-step tool dependencies and evolving task context. This limitation often introduces irrelevant tools that mislead the agent, degrading efficiency and accuracy. We propose Dynamic Tool Dependency Retrieval (DTDR), a lightweight retrieval method that conditions on both the initial query and the evolving tool calling plan. DTDR models tool dependencies from function calling demonstrations, enabling adaptive retrieval as plans unfold. We benchmark DTDR against state-of-the-art retrieval methods across multiple datasets and LLM backbones, evaluating retrieval precision, downstream task accuracy, and computational efficiency. Additionally, we explore strategies to integrate retrieved tools into prompts. Our results show that DTDR improves function calling success rates between $23\%$ and $104\%$ compared to state-of-the-art static retrievers. 
\end{abstract}

\section{Introduction}

\begin{figure*}[ht]
    \centering
    \includegraphics[width=\textwidth]{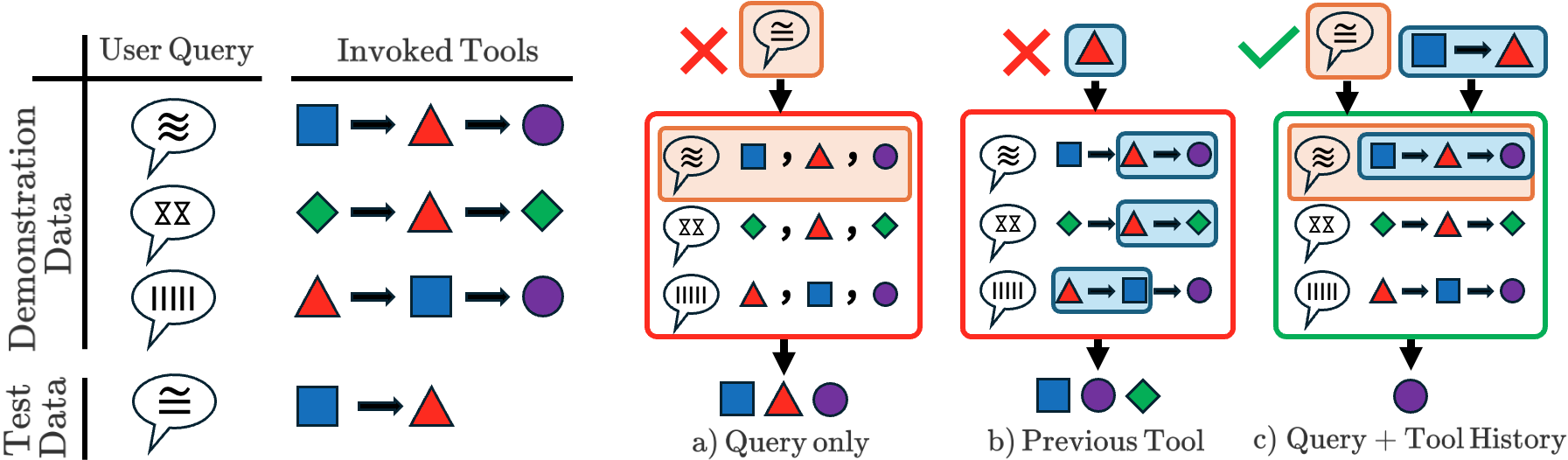}
    \caption{\textbf{Dynamic Tool Dependency Retrieval.} Given demonstration data for a set of tools, previous work retrieves tools based on either a) the natural language query (highlighted in orange) %\textcolor{orange}{orange}
    or b) the latest executed tool call in the plan (the red triangle, highlighted in teal). %\textcolor{tealblue}{teal}
    We instead propose a retrieval method that is conditioned on both the query and the growing history of tool calls (blue square, then red triangle). This method allows for retrieving only tools that are strictly relevant to both query (orange%\textcolor{orange}{orange}
    ) and tool context (teal%\textcolor{tealblue}{teal}
    ).}
    \label{fig:fig1}
\end{figure*}

% \footnotetext[*]{Equal Contribution.}
\begingroup
\renewcommand{\thefootnote}{\fnsymbol{footnote}}
\footnotetext[1]{Equal Contribution.}
\endgroup
\footnotetext[1]{Qualcomm AI Research is an initiative of Qualcomm Technologies, Inc.}
\footnotetext[2]{Work done during an internship at Qualcomm AI Research, Amsterdam.}

Large language models (LLMs) augmented with tool use (a.k.a., function calling) have rapidly evolved from early neuro-symbolic systems to agentic frameworks that can plan, select, and invoke external Application Programming Interfaces (APIs) \citep{yao2023react,schick2023toolformer,patil2024gorilla, patel2025learning}. 
Despite this progress, deploying tool-augmented LLMs on-device remains challenging due to two key constraints: (i) \emph{lighweight} under strict memory and latency budgets, and (ii) \emph{effectiveness} across large and heterogeneous tool sets. 

Therefore, prior work has proposed using tool \textit{retrieval} modules to encode only relevant tools into the prompt of the function calling agent \citep{qin2023toolllm,braunschweiler2025toolreagt,paramanayakam2025less}. Doing so makes it easier for the agent to identify the correct function, reducing unnecessary calls, and enhancing both accuracy and prompt efficiency. However, a major challenge is determining what information should guide the tool retrieval. Some methods rely solely on semantic similarity between the query and tool descriptions \citep{gao2025tool, paramanayakam2025less}, which can overlook the history of selected tools in the ongoing, multi-step plan. Others leverage static tool dependency graphs built from demonstration trajectories \citep{liu2024toolnet}, but these approaches risk retrieving tools irrelevant to the query or being biased toward repeated calls.  

An effective tool retriever must adapt to both the \emph{current task} and the \emph{ongoing trajectory}, while remaining lightweight enough to satisfy on-device constraints. This setting raises the question: can such specificity be achieved with a low-resource approach? To answer this, we introduce \textbf{Dynamic Tool Dependency Retrieval (\ours)}, a lightweight retrieval component that, given a user query and partial plan, identifies a small set of relevant tools and their dependency relations (see Figure \ref{fig:fig1}). Our main contributions are as follows:

% Recent work \citep{liu2024toolnet, gao2025tool, chen2025gtool} has proposed using tool dependencies to further aid the LLM in selecting appropriate function calls.

% With dependency-aware tool retrieval, on-device agents save prompt tokens and avoid execution calls on irrelevant tools, improving end-to-end accuracy, latency, and energy efficiency. However prior w

\begin{enumerate}[left=0pt]
\item \textbf{Lightweight Tool Dependency Retrieval method.} We formulate \emph{Dynamic Tool Dependency Retrieval} (\ours), a dependency-aware tool retrieval framework that conditions on both the user query and the evolving tool plan to recover a minimal, task-specific dependency subgraph.  Through comprehensive analysis, we provide strong evidence that methods without query- and history-awareness are unable to solve the tool retrieval task, so history-aware methods should be adopted.
\item \textbf{Extensive Retrieval and End-to-End evaluation.} We conduct a systematic comparison against a suite of retrieval baselines (text-based, embeddings-based, and dependency/graph-based), showing that our dynamic variant outperforms previous work in terms of \emph{retrieval} metrics (MRR/$F_1$-score), \emph{downstream} performance (function selection accuracy and end-to-end task success), and \emph{efficiency} (footprint, token budget) across several datasets and LLM backbones of varying sizes. 
\item \textbf{Analysis with multiple Prompting strategies.}  We use a prompt-efficient in-context learning (ICL) representation that conditions the LLM only on the minimal subgraph of tool dependencies retrieved. We benchmark multiple ICL encoding strategies, identifying weighted Hard Masking as the best contender, but also investigating when and why other approaches should be preferred based on different factors (model scale, dataset statistics, tool retrieval accuracy).
\end{enumerate}

\section{Related Work}\label{sec:related_work}
% \todo{Proposal, also mentioned by Bence, is to somehow rename query-aware and history-aware to something about being dynamic, since that is the main thing we are claiming in our work} \todo{Explain this. why is it good to be tool dependency aware but bad to be tool description aware?}

\begin{table*}[tbp]
\centering
\resizebox{\textwidth}{!}{%
\begin{tabular}{@{}lcccccccc@{}}
\toprule
\textbf{Method} & \makecell{\textbf{Tool Deps.} \\ \textbf{Aware}} & \makecell{\textbf{Tool Desc.} \\ \textbf{Free}}  & \makecell{\textbf{Query} \\ \textbf{Aware}} & \makecell{\textbf{Multi-Step} \\ \textbf{History Aware}} & \makecell{\textbf{Small} \\ \textbf{model}} \\
\midrule
BM25 \citep{robertson2009probabilistic} & \xmark & \xmark & \cmark & \xmark  & \cmark \\
ToolGraph Retriever \citep{gao2025tool} & \cmark & \xmark & \cmark & \xmark   & \xmark \\
Less-is-More Lv1 \citep{paramanayakam2025less} & \xmark & \xmark & \cmark & \xmark  & \cmark \\
ToolNet \citep{liu2024toolnet}  & \cmark & \cmark & \xmark &  \xmark  &  \cmark \\
TinyAgent \citep{erdogan2024tinyagent}& \xmark& \cmark & \cmark & \xmark  & \cmark \\
% ControlLLM \citep{liu2024controlllm}& \cmark& \xmark & \cmark & \cmark  & \xmark \\
Toolformer \citep{schick2023toolformer}& \xmark & \cmark & \cmark & \xmark  & \cmark \\ 
\textbf{\ours (Ours)} & \cmark & \cmark & \cmark & \cmark & \cmark \\
\bottomrule
\end{tabular}
}
\caption{Comparison of tool retrieval methods used in prior work. Our work is the only one that satisfies all $5$ desired conditions.}
\label{tab:tool_comparison}
\end{table*}

Recent LLMs demonstrate impressive tool-usage capabilities, with cloud-based models such as GPT-5 \citep{openai2025gpt5systemcard}, Claude 4 \citep{anthropic2025claude4systemcard}, and GLM-4.5 \citep{zeng2025glm} leading by a wide margin on function-call benchmarks like BFCL V4 \citep{patilberkeley} and $\tau^2$-Bench \citep{barres2025tau}. Recent work has explored fine-tuning LLMs on large datasets to build general-purpose function calling models \citep{gorilla-openfunctions-v2}. These models often fail in real-world scenarios due to poor function selection, misinterpretation of user intent, or under realistic data perturbations \citep{dang2025improving,rabinovich2025robustness}. Alternatively, ICL strategies for tool learning can involve including tool descriptions \citep{shen2023hugginggpt, shen2024taskbench, patel2025learning} or example trajectories \citep{paranjape2023art,sarukkai2025self} within the model prompt. To handle hundreds of functions, recent work uses retrieval-augmented generation to sub-select tools that are relevant to the task \citep{qin2023toolllm, braunschweiler2025toolreagt, paramanayakam2025less}. Most prior art retrieves relevant tools based on the query and tool descriptions \citep{braunschweiler2025toolreagt,paramanayakam2025less,paranjape2023art}. \cite{liu2024controlllm} and \cite{ding2025scitoolagent} assume a known tool dependency graph and use it to aid the LLM with selecting relevant functions. As tool dependencies are often unknown, others propose learning them via demonstrations \citep{paranjape2023art, chen2025gtool, liu2024toolnet, qin2023toolllm, patil2024gorilla, erdogan2024tinyagent}. Unlike prior retrievers that rely solely on the user query or static tool dependency graphs, we propose a dynamic tool retrieval approach that conditions on both the current query and the evolving sequence of previously invoked tools. 
In Table~\ref{tab:tool_comparison}, we summarize related works that most closely align with our method. We compare against $5$ different categories. \textbf{1) Tool Dependency Aware:} retrievers that utilize tool dependencies are better informed to solve multi-step tasks \citep{gao2025tool}; \textbf{2) Tool Description Free:} function documentation may not be available or consistent across large sets of tools \citep{patel2025learning}; \textbf{3) Query Aware:} retrieval methods should retrieve tools relevant to the specific task, i.e. email tasks will more likely need getting an email address than opening a document, \textbf{4) Multi-Step History Aware:} retrievers that take into account multi-step history, not just current step, avoid bias towards frequently called functions,  and \textbf{5) Small Model:} On-device agents have stricter memory and latency requirements that the retrieval method needs to satisfy. Our method is the only one that satisfies all $5$ categories.
A more extensive discussion of related works can be found in Appendix~\ref{sec:app_related_works}.

\section{Problem Formulation}\label{sec:prob_form}

\paragraph{Retrieval-Augmented Function Selection.} Consider the function selection agent $\pi$, consisting of a frozen language model (LM). 
%Because the agent consists of only the single LM-based policy $\pi$, we will use the terms ``policy'' and ``agent'' interchangeably, so $\pi$ refers to both. 
Let $\funcF$ be the collection of function names that the agent can choose from to solve a task described through a natural language query $q$, such as \textit{``Reply to my latest email from Willem.''}.  Let $f_{0:t-1} = \left[f_0, f_1, ..., f_{t-1} \right]$ be the list of previous functions predicted for $q$ up until timestep $t$. Typically, $q$, $f_{0:t-1}$ and $\funcF$ are encoded in the LM prompt, $p$, which is input to the agent to sample a function $f_t \sim \pi(\cdot | p)$. A sampled function $f_t$ is correct if $f_t \in \funcF_{t}^*$, the set of correct functions for $q$ at $t$. To raise the probability of said correctness, prior work encodes into the prompt a subset of $\funcF$ \citep{qin2023toolllm,braunschweiler2025toolreagt}. This subset is retrieved via some module $\omega(\cdot)$, a function that outputs a subset of function names $\funcF_t \subseteq \funcF$. The set of inputs to the $\omega$ is dependent on the specific retrieval method. Let $T$ be the maximum trajectory length for any $q$, let $\Omega$ be the set of all possible $\omega$ retriever functions, and let $\mathbf 1$ be a scalar indicator function. Our function selection problem is a prompt optimization objective: 
\begin{equation}
  \max_{\omega \in \Omega} \sum_{q \in Q} \sum_{t=0}^T \mathbf E_{f_t \sim \pi(\cdot|p)}[\mathbf 1_{f_t \in \funcF_{t}^*}].  
\end{equation}

% \begin{equation}
% \max_{\omega \in \Omega} \sum_{q \in Q} \sum_{t=0}^T \mathbf{E}_{f_t \sim \pi(\cdot|p(q, f_{0:t-1}, \funcF_t)) }[\mathbf 1_{f_t \in \funcF_{t}^*}]
% \end{equation}
% Each function $f \in \funcF$ consists of both a name, $f_{name}$, and a description, $f_{desc}$. For notation simplicity, unless otherwise stated $f = f_{name}$.

% \begin{equation}\label{eq:selection_maximization}
%     \max_{\omega \in \Omega} \sum_{q \in Q} \sum_{t=0}^T \mathbf E_{f_t \sim \pi(\cdot|p(q, f_{0:t-1}, \funcF_t)) }[\mathbf 1_{f_t \in \funcF_{t}^*}].
% \end{equation}
This optimization problem mathematically grounds the success of the function calling agent $\pi$ to how well the retrieved set $\funcF_t$ aligns with $\funcF_{t}^*$. Specifically, we consider an optimal retrieval module $\omega^*$ as one such that it retrieves a non-empty subset of the ground-truth set: $\{\} \subset \omega^*(\cdot) = \funcF_{t} \subseteq \funcF_{t}^*$. Note that we focus on optimizing the individual function selection step for more fine-grained downstream evaluation of the retrieval module, so this objective function is less sparse than ones for multi-step, function-calling tasks \citep{patel2025learning}. We discuss in Section \ref{sec:results} end-to-end results with trajectory-level evaluation.

In our setting, the agent does the entire planning (retrieval and function selection) without any function calling interleaved. Our scenario for tool selection remains a valid and practical setting for agentic pipelines. In many real-world systems—such as AI-based IDE assistants \citep{antigravity2025}, workflow automation agents, or enterprise copilots—the agent first generates a plan and only then executes the selected tools. In these settings, accurate tool selection at planning time is critical: incorrect planning can lead to unnecessary execution and costly re-planning. Therefore, improving tool selection in an execution-free, offline setting directly strengthens the planning phase of agentic systems and increases efficiency and latency predictability before any tool invocation occurs.

% Note that we treat $\pi$ as probabilistic in Equation \ref{eq:selection_maximization} due to the inherent non-determinism of SoTA language models. 

\begin{figure*}[!h]
    \centering
    \includegraphics[width=0.8\textwidth]{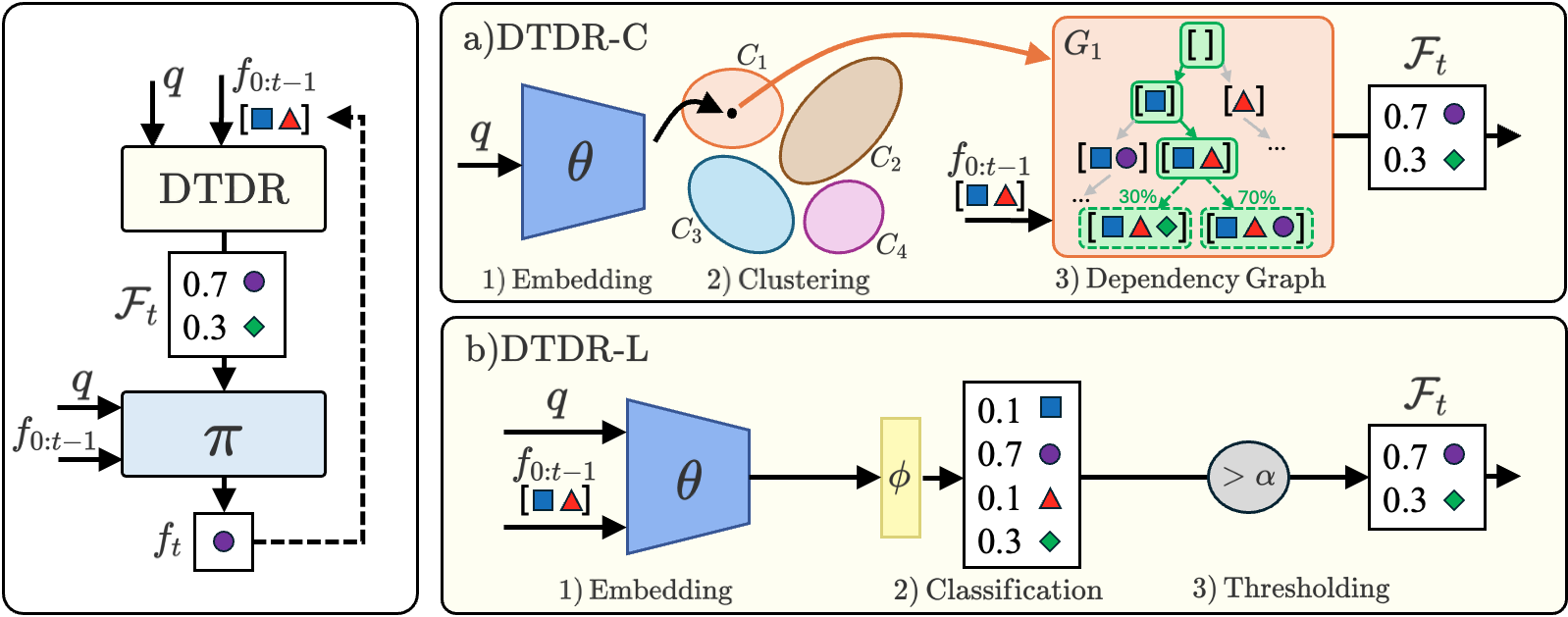}
    \caption{\textbf{System diagram for \ours}. On the left, the user query and tool history are input to DTDR to retrieve the most likely next tools. The LLM $\pi$ selects the next tool among this set. On the right, we show the two alternative instantiations for the retriever: a) \ourscl, based on a clustering step to retrieve an explicit graph of tool dependencies; and b) \oursnn, based on a learned linear classifier implicitly modeling tool dependencies. Both systems are conditioned on user query and full tool history.}

    \label{fig:system}
\end{figure*}

\paragraph{Limitation of Prior Work: Suboptimal Sets of Inputs.} Some zero-shot retrieval methods \citep{paramanayakam2025less, gao2025tool} implement $\omega$ as the cosine similarity between embeddings of the query (or a transformation of it) and embeddings of each tool description. Furthermore, the retrieved set $\funcF_t$ is used throughout the entire trajectory, constant throughout the $T$ timesteps. Relying solely on static tool descriptions often overlook the dynamic nature of tool usage in context. While semantic similarity between queries and tool descriptions may capture surface-level relevance \citep{lin2024hammer}, it fails to account for how tools are actually sequenced in real tasks. Other work such as \cite{liu2024toolnet} utilize collected demonstration trajectory data $D$ of function calling tasks, obtained by tool providers \citep{paranjape2023art}, historical tool trajectories \citep{chen2025gtool} or generated \citep{erdogan2024tinyagent, gao2025tool, patil2024gorilla}. With $D$, they compute empirical next-function probabilities based on the latest function $f_{t-1}$ to obtain $\funcF_t$. 
However, this limited context does not model multi-step tool dependencies well. For example, if sending the same email to multiple people, \textit{getEmailAddress} must be called consecutively multiple times. Demonstrations that contain function trajectories such as these will cause a high value of $P$(\text{\textit{getEmailAddress}}|\text{\textit{getEmailAddress}}), thus biasing the agent to repeatedly call it more times than necessary. Furthermore, computing probabilities based on the entire demonstration data across various queries underfits tool-use patterns. The $P(\text{\textit{getEmailAddress}}|\text{\textit{getEmailAddress}})$ value suitable in tasks with multiple recipients can be misleading in tasks with a single recipient. More generally, conditioning only on the most recent function fails to capture branching dependencies that depend jointly on earlier steps and task intent. For instance, after calling \textit{open\_file} followed by \textit{summarize\_pdf}, the appropriate next action differs depending on whether the task requires appending notes, composing an email, or performing another file operation. A first-order model aggregates all continuations of \textit{summarize\_pdf} across diverse tasks, assigning non-zero probabilities to tools that are valid in other queries but not appropriate in the current plan. This misalignment results in an overly broad candidate set that lacks task-specific precision. Without incorporating task-specific or higher-order information, the context of previous functions can therefore act as a misleading prior for $\pi$ Without the task-specific information, the context of previous functions may be a misleading prior for $\pi$. We aim for a \textit{lightweight retrieval mechanism to obtain tool dependency priors that capture both task-specific and context-specific information}.

\section{Proposed Approach: Dynamic Tool Dependency Retrieval}

To capture appropriate tool dependencies for Retrieval-Augmented Function Calling, we introduce \textit{Dynamic Tool Dependency Retrieval (DTDR)}. The high-level idea of DTDR is simple: the retrieval module $\omega$ should be based on both the task $q$ and the trajectory of previous function calls $f_{0: t-1}$; thus, $\omega(q, f_{0: t-1})$. Figure~\ref{fig:system} (left) gives an overview of our approach. The test query $q$ and the trajectory of function calls $f_{0: t-1}$ are input to \ours, which predicts a set of tools $\mathcal{F}_{t}$, assigning a probability to each. We then encode the retrieved dependencies into the prompt $p$ by hard masking the full set of tools $\mathcal{F}$, while also providing the probabilities for the retrieved tools. In Section \ref{sec:exp_setup}, we elaborate on other ICL methods we investigated to embed $\funcF_t$ into the prompt. The encoded prompt is then processed by the LLM agent to sample the next function $f_{t} \sim \pi(\cdot | p)$. For end-to-end evaluation, this continues in an iterative approach starting from the empty function call plan, and ending after predicting the ``end-of-plan'' function.

%\bhrij{We need to mention somewhere in the paper, either as a main part of our story or experiment detail about why we do not do online updating of tool dependency graph/tool dependency graph learning. online automatic updating of tool learning from agent experience is popular in literature (though not practical). }

\paragraph{Variants of DTDR.} We propose two lightweight variants for DTDR: one supervised gradient-based method and one unsupervised clustering-based method. Firstly, proposing two allows us to validate whether our claims on \emph{dynamic} tool retrieval are consistent across different categories of methods. Secondly, it enables us to directly compare against previous work belonging to these two categories. Lastly, depending on software and system constraints, either variant might be preferred for a real on-device scenario. Both variants utilize demonstration data $\mathcal{D}$. 
%We focus on the offline setting rather than automatic updating as offline learning allows for more controlled updates in sensitive environments such as mobile phones.

\begin{table*}[!h]
\centering
\resizebox{0.9\textwidth}{!}{%
\begin{tabular}{l|ccc|ccc|ccc}
\toprule
% &  & \multirowcell{3}{\textbf{Fun.} \textbf{Sel.}\\\textbf{Acc. [\%] ($\uparrow$)}} & & \\
%  & & & \multicolumn{2}{c}{\textbf{Retrieval Metrics}} \\
% & \textbf{Retrieval Method} &  \textbf{FSA [\%] (\uparrow)} & ~~\textbf{MRR} \textbf{($\uparrow$)}~~ & ~~\textbf{F$_1$} \textbf{($\uparrow$)}~~ \\ % & \textbf{FLOPS ($\downarrow$)} \\
\multirowcell{2}{\textbf{Retrieval Method}} & \multicolumn{3}{c}{\textbf{TinyAgent}} & \multicolumn{3}{c}{\textbf{Taskbench-HF}} & \multicolumn{3}{c}{\textbf{Taskbench-MM}} \\

& \textbf{FSA}  &\textbf{ MRR }&\textbf{ F$_1$} &  \textbf{FSA} & \textbf{MRR} & \textbf{ F$_1$} &  \textbf{FSA} & \textbf{MRR} & \textbf{ F$_1$}
% &  \textbf{FSA [\%] (\uparrow)} & \textbf{MRR} \textbf{($\uparrow$)} & \textbf{F$_1$} \textbf{($\uparrow$)} &  \textbf{FSA [\%] (\uparrow)} & \textbf{MRR} \textbf{($\uparrow$)} & \textbf{F$_1$} \textbf{($\uparrow$)}
\\
\midrule

  Random Guess & \05.9& 0.26& 0.12 & \04.2 & 0.16 & 0.05  & \02.4 & 0.11& 0.03\\
  BM-25 \citep{robertson2009probabilistic} & 23.1 & 0.35 & 0.20 & \08.3 & 0.18 & 0.05 & 13.7 & 0.26 &	0.19 \\
  QTS (Vanilla) & 15.8 &	0.26 &	0.14 & \07.9 & 0.21 & 0.08 & \05.3	&0.14	&0.04\\
  QTS \citep{gao2025tool}  & 21.5&	0.18 &	0.09 & 14.6 & 0.37 & 0.27 &  \05.3	&0.14	&0.20\\
  QTS \citep{paramanayakam2025less} & 23.7 &	0.36 &	0.19 & 12.9 & 0.30 & 0.18  & \07.6 &	0.29	&	0.18\\
  DR \citep{liu2024toolnet} & 30.7 &	0.70& 0.49 & 17.1 & 0.46 & 0.33 & 13.3	&0.38	&0.27	\\
  \cellcolor{verylightgray}Dynamic DR (\ourscl) (Ours) & \cellcolor{verylightgray}43.3&	\cellcolor{verylightgray}0.78 &	\cellcolor{verylightgray}\bfu{0.56} & \cellcolor{verylightgray}27.5& \cellcolor{verylightgray}0.64 & \cellcolor{verylightgray}0.55 &  \cellcolor{verylightgray}24.4	&\cellcolor{verylightgray}0.52	&\cellcolor{verylightgray}0.43\\
  LR \citep{erdogan2024tinyagent} & 25.6	 &0.53&	0.39 & 20.5 & 0.53 & 0.32 &  21.0	&0.49 & 0.28\\
  \cellcolor{verylightgray}Dynamic LR (\oursnn) (Ours)  & \cellcolor{verylightgray}\bfu{65.1}&	\cellcolor{verylightgray}\bfu{0.93} &	\cellcolor{verylightgray}0.55 & \cellcolor{verylightgray}\bfu{27.8} & \cellcolor{verylightgray}\bfu{0.75} & \cellcolor{verylightgray}\bfu{0.63}&  \cellcolor{verylightgray}\bfu{27.0}&	\cellcolor{verylightgray}\bfu{0.69}	&\cellcolor{verylightgray}\bfu{0.55}\\

\bottomrule
\end{tabular}
}
\caption{Retrieval performance for methods from different categories (QTS = Query-Tool Similarity, LR = Learned Retriever, DR = Dependency Retriever). QTS and LR categories are query-conditioned, while DR is history-conditioned. Our Dynamic methods are conditioned on both query and tool history, which yields improved performance. Function Selection Accuracy is reported on Qwen 3 0.6B using hard masking as ICL method. Best results per dataset and metrics are in bold.}
\label{table:retrieval_Amir}
\end{table*}

\paragraph{Dynamic Tool Dependency Retrieval-Clustering (\ourscl).} This variant (see Figure~\ref{fig:system}, top-right) is based on a clustering component and a graph-traversal component. The clustering component maps a test query to a tool dependency graph which is relevant for the specific task. The graph-traversal component traverses the graph based on the history of tool calls, and determines the most likely next tools for the plan.
Formally, we embed demonstration queries $Q_{\mathcal{D}}$ with a pretrained embedding model and fit a $K$-Means clustering model $C$ on the embeddings. Let $\beta$ be the embedding of $q$ and $d$ be the ground-truth demonstration data for $q$. For each cluster index $k$, we consider the set of demonstrations assigned to that cluster $D_{k} = \{ d \mid \forall q \in Q_{\mathcal{D}} \; s.t. \ C(\beta) = k \}$. We build a weighted tool dependency graph $G_{k}$ describing the next-tool dependency probabilities given a sequence of tools. In particular, $G_{k}(f_{0:t-1})$ describes the set of functions which follow the history $f_{0:t-1}$ in the demonstration data $D_{k}$, as well as their probabilities. Details on constructing the tool dependency graph $G$ are included in Appendix ~\ref{app:implementation} and Appendix~\ref{app:pseudocode}.
Given a test query $q$ and history $f_{0:t-1}$, the next-tool prediction in \ourscl can be obtained with  $\funcF_t = G_{C(\beta)}(f_{0:t-1})$. The number of learned parameters in the $K$-means model is $e * K$, with $e$ being the embedding dimensionality and $K$ the number of clusters.

\paragraph{Dynamic Tool Dependency Retrieval-Linear (\oursnn).} This supervised learning variant (see Figure~\ref{fig:system}, bottom-right) is based on a linear layer classifier trained to predict the set of next functions given the test query and the history of previous tool calls. We train a $1$-linear-layer classifier $\phi$ on top of a frozen embedding model to predict the next function given both the query and the function trajectory until $t$. Let $\zeta$ be the embedding of $q + f_{0:t-1}$,  where the operator $+$ denotes concatenation between strings. We use a softmax operation to obtain a probability distribution from the model outputs. Thus $\phi(\zeta)$ is the probability of $f$ given the query and ongoing, current trajectory. To retrieve only a subset of tools, we set a threshold value $\alpha$, so that $\funcF_t  = \{f | f \in \funcF, \phi(\zeta) > \alpha \}$. The number of learned parameters in this model is $e * |\funcF|$, only depending on the output size of the embedding model $e$ and number of tools $|\funcF|$. Additional implementation details and pseudocode can be found in Appendix~\ref{app:implementation} and Appendix~\ref{app:pseudocode}.

\section{Experimental Setup}\label{sec:exp_setup}

\paragraph{Datasets.}
We benchmark on four function calling datasets: \emph{TinyAgent} \citep{erdogan2024tinyagent}, \emph{TaskBench DailyLife APIs}, \emph{TaskBench HuggingFace}, and \emph{TaskBench Multimedia} \citep{shen2024taskbench}. The first two datasets represent tools commonly available on a typical device, the other two datasets are specific to a particular domain, which makes them more challenging for ``out-of-the-box'' LLMs. We report in Appendix~\ref{app:datasets} the statistics for each benchmark, with an important insight being that all datasets have between 1 and 2 tool dependencies per plan, except for TaskBench DailyLife APIs which has almost zero dependencies. For each dataset, we set aside around 30\% of the data for testing, while the rest is used as demonstration trajectories for tool retrieval and ICL methods. 
%In Figure~\ref{fig:ablations}c, we investigate decreasing the number of demonstrations.

\paragraph{LLM backbones.}
We evaluate tool-retrieval and ICL methods applied to $7$ LLM backbones. We consider the \emph{Qwen 3} family of models (0.6B, 1.7B, 4B, 8B, 14B) \citep{yang2025qwen3} representing edge devices of varying size, \emph{GPT-4o} \citep{hurst2024gpt} representing a cloud model solution, and \emph{Gorilla-V2} \citep{gorilla-openfunctions-v2} representing an edge-device fine-tuned on function calling data. Assuming INT4 quantization and a KV cache of up to 10 thousand tokens, Qwen 3 models up to Qwen 3 4B could efficiently run on a typical mobile device \citep{federiciefficient,song2025harnessing}.

\paragraph{Metrics.} For retrieval performance, we measure \emph{Mean Reciprocal Rank (MRR)} and  $\mathit{F_1}$ score. To analyze downstream performance, we look at both \emph{Function Selection Accuracy (FSA)} and \emph{Success Rate (SR)} for the end-to-end task, which considers both function selection \textbf{and} parameter filling. Full details in Appendix \ref{sec:metric_details}. We assess efficiency via \emph{prompt length} as a proxy for prefill speed (or time to encode and process the whole prompt); and \emph{number of parameters} to quantify the LLM footprint which determines on-device usability.

\paragraph{Tool Retrieval Baselines.}
We consider different categories of Tool Retrievals baselines from recent work presented in Table~\ref{tab:tool_comparison}. Classical term-similarity baselines like BM-25 \citep{robertson2009probabilistic} directly compare text similarity between documents. \emph{Query-Tool Similarity (QTS)} baselines use pre-trained sentence-embedding models to encode descriptions of queries and tools, without learning from demonstrations. This approach is extended in Less-is-More Level 1 \citep{paramanayakam2025less} by prompting an LLM to describe the ideal tool set to solve a query, and in Tool Graph Retriever \citep{gao2025tool} by additionally encoding tool dependencies in the embeddings. \emph{Learned Retriever (LR)} baselines as in \citet{erdogan2024tinyagent} fine-tune small classifiers to learn what functions are useful to resolve a query. Finally, \emph{Dependency Retriever (DR)} methods like ToolNet \citep{liu2024toolnet} leverage tool dependencies to directly sub-select viable tools based on the most recent tool in the plan.
%In contrast, \ours dynamically retrieves relevant tool dependencies based on the query \emph{and} the history of tool calls in the plan. We propose a \emph{Dynamic Learned Retriever} (\oursnn), which jointly processes the query and the history to predict the next tool. We also propose a \emph{Dynamic Dependency Retriever} (\ourscl), which clusters demonstrations based on their embeddings, and builds a higher-order Markov Chain dependency graph for each cluster, dynamically retrieved based on the query and the history. 
\ours introduces \emph{dynamic} retrieval methods that leverage both the \emph{query} and evolving \emph{tool-call history} to improve tool prediction. These are referred to as a \emph{Dynamic Learned Retriever} (\oursnn) and a \emph{Dynamic Dependency Retriever} (\ourscl). As with \ours, our baselines do not assume execution feedback like works such as SEER \citep{cui-etal-2025-self}.
Additional details for baselines and methods in Appendix \ref{app:implementation}.

% We compare our DTDR variants to different categories of tool retrieval baselines. 
% \begin{itemize}
%     \item Random 
%     \item Query Tool-Description Similarity (QTS): We apply
% \end{itemize}
% \textbf{BM-25} \citep{robertson2009probabilistic} 

% Methods: \\
% TF-IDF \citep{sparck1972statistical}  \\
% BM-25 \citep{robertson2009probabilistic} \\
% ToolNet \citep{liu2024toolnet} \\
% Query-Tool Retrieval \\
% ToolGR \citep{gao2025tool} \\
% Less-is-More Lv1 \citep{paramanayakam2025less} 

\paragraph{ICL Methods for Encoding $\mathcal F_t$ into Prompt.} For all methods, including the \emph{No ICL} baseline, we provide 1 randomly selected demonstration to show how to complete an example task. For all other ICL methods, we encode the retrieved tools. For the \emph{Raw Demonstrations} method, we provide up to $5$ additional demonstrations for which the plan includes the predicted tools. For more efficient prompting, we can either use the \emph{Hard Mask} and completely exclude the remaining tools from the prompt, or a \emph{Soft Mask}, where the full set of tools is presented to the model but the retrieved list is emphasized as the set the model should generally prefer. Since retrieval methods provide scores for each tool, these can be included encoded, yielding \emph{Weighted} variants of the two masking approaches. We provide example ICL prompts in Appendix \ref{app:prompt}.

\section{Results and Discussion}\label{sec:results}

\begin{figure*}[t]
    \centering
    \includegraphics[width=\textwidth]{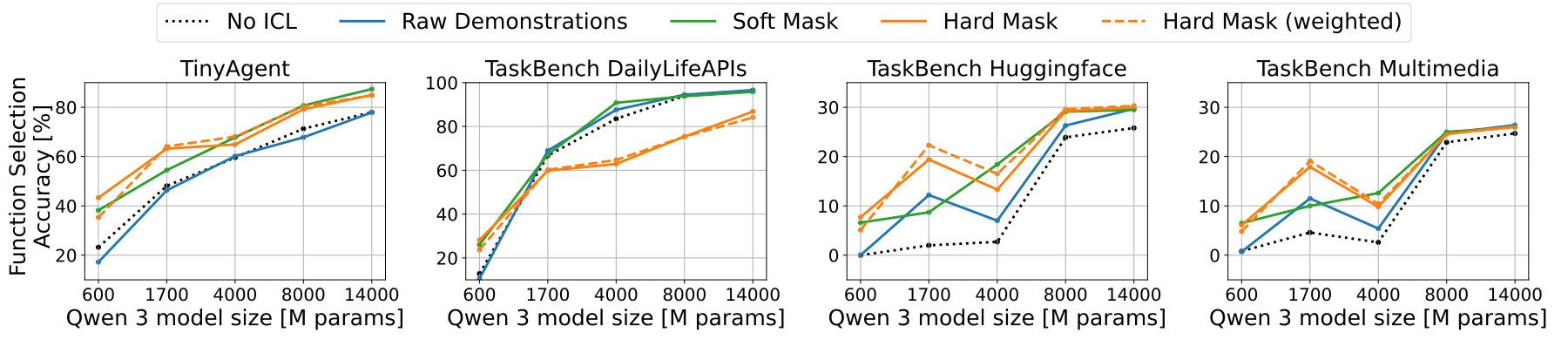}
    \caption{Comparison of efficient ICL methods against ICL with Raw Demonstrations and the baseline without ICL. All results are conditioned using the \ourscl retriever. Pruning irrelevant tools from the prompt has a greater impact for smaller models that cannot handle longer contexts.}
 
    \label{fig:icl}
  
\end{figure*}

\begin{figure}
    \centering
    \includegraphics[width=\linewidth]{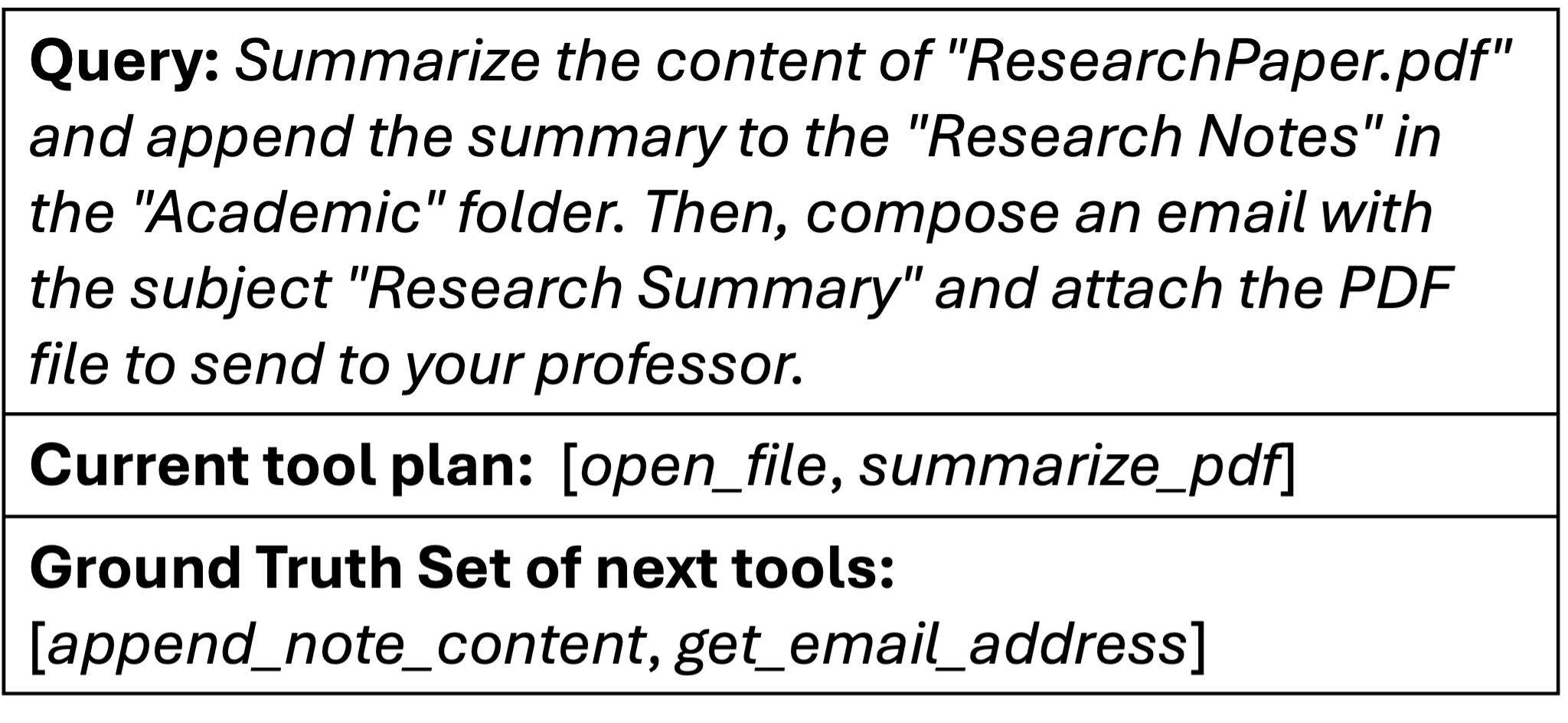}
    \caption{Example tool selection step in TinyAgent.}
    \label{fig:qual_example}
\end{figure}

\paragraph{Does DTDR improve tool retrieval?} We evaluate Tool Retrievers on ranking, retrieval and downstream performance (see Table \ref{table:retrieval_Amir}). MRR scores generally correlate well with Function Selection Accuracy, indicating that better ranking improves LLM reasoning. Set-based metrics like $F_1$ score are less informative of the downstream performance, as they do not capture the relative importance of tools. Overall, \emph{Learned Retriever} and \emph{Dependency Retriever} perform better than all baselines. Our \emph{dynamic} Learned Retriever (\oursnn) surpasses the \emph{static} counterpart \citep{erdogan2024tinyagent} by over 35\% on TinyAgent and TaskBench DL, while matching performance on TaskBench HF and MM. This demonstrates the benefit of conditioning the prediction on the function selection history. Similarly, \ourscl improves over the \emph{static} Dependency Retriever counterpart \citep{liu2024toolnet} by 50--100\%, confirming the value of leveraging query and full history for retrieval, instead of only considering the latest function in the plan. For a qualitative analysis, we compare tools retrieved from DR, LR, and DTDR-L for the example in Figure~\ref{fig:qual_example}. For DR, the last tool call in the history (\textit{summarize\_pdf}) is the only input for this tool retriever. As \textit{summarize\_pdf} appears in queries with different contexts in the demonstration data, DR predicts a high probability (between 5\% and 20\%) for $7$ tools. Because many TinyAgent queries involving PDFs require opening and summarizing files, LR (query-aware only) assigns high probability to [\textit{summarize\_pdf, open\_file, end\_plan}], even though these tools have already been selected in the current plan. DTDR-L retrieves [\textit{append\_note\_content, get\_email\_address, compose\_email}]. 

\begin{table*}[!h]
\centering
\resizebox{0.9\textwidth}{!}{%
\begin{tabular}{l|l|cc|cc|cc|cc}
\toprule
 % \multicolumn{6}{c}{\textbf{Function Selection Accuracy [\%] ($\uparrow$)}}\\
%\midrule

    % & \textbf{ICL Method} & \multicolumn{4}{c}{\textbf{Function Selection Accuracy ($\uparrow$)}} \\
    
 % & \multirowcell{2}{\textbf{ICL Method}} & \multicolumn{2}{|c|}{TinyAgent} & \multicolumn{2}{|c|}{TaskBench \\ DailyLife}  \\

 % \multirowcell{2}{\textbf{Model}} & \multirowcell{2}{\textbf{Method}} & \multicolumn{2}{c}{\textbf{TinyAgent}} & \multicolumn{2}{|c|}{\textbf{TB-Dailylife}}
  \multirowcell{2}{} & \multirowcell{2}{\textbf{Method}} & \multicolumn{2}{c|}{\textbf{TinyAgent}} & \multicolumn{2}{c}{\textbf{TB-DL}} & \multicolumn{2}{c}{\textbf{TB-HF}} & \multicolumn{2}{c}{\textbf{TB-MM}} 
 \\
 & & FSA ($\uparrow$) & SR ($\uparrow$) & FSA ($\uparrow$) & SR ($\uparrow$) & FSA ($\uparrow$) & SR ($\uparrow$) & FSA ($\uparrow$) & SR ($\uparrow$)
 \\
\midrule
\multirow{5}{*}{\rotatebox[origin=c]{90}{Qwen3 0.6B~~}}
 & No ICL                                   & 23.2 & \00.4 & 12.8 &\00.0 & \05.5 & \00.0 & \07.3 & \00.0 \\
 % & Raw Demonstrations       & 17.2 & \00.1 & 10.5 & \00.0 & \04.3 & \00.0 & \05.3 & \00.0 \\
 & DR \citep{liu2024toolnet}       & 22.0 & \00.2 & 11.5 & \00.0 & \06.7 & \00.0 & \09.5 & \00.3 \\
 & LR \citep{erdogan2024tinyagent}   & 19.8 & \00.9 & 19.6 & \02.0 & 10.0 & \00.5 & 12.1 &  \00.7 \\
 & \cellcolor{verylightgray}\ourscl (Ours)  & \cellcolor{verylightgray}35.3 & \cellcolor{verylightgray}\0\bfu{4.2} & \cellcolor{verylightgray}23.7 & \cellcolor{verylightgray}\0\bfu{6.8} & \cellcolor{verylightgray}\bfu{20.8} & \cellcolor{verylightgray}\0\bfu{5.9}  & \cellcolor{verylightgray}\bfu{20.2} & \cellcolor{verylightgray}\0\bfu{8.3}\\
 & \cellcolor{verylightgray}\oursnn (Ours)       & \cellcolor{verylightgray}\bfu{46.1} &\cellcolor{verylightgray} \03.5 & \cellcolor{verylightgray}\bfu{45.9} & \cellcolor{verylightgray}\00.8 & \cellcolor{verylightgray}18.9 & \cellcolor{verylightgray}\01.2 & \cellcolor{verylightgray}17.6 & \cellcolor{verylightgray}\03.7 \\
\midrule
\multirow{5}{*}{\rotatebox[origin=c]{90}{Qwen3 1.7B~~}}
 & No ICL                                   & 48.1 & \04.4 & 66.7 & 13.5 & 32.2 & \03.6 & 38.1 & \04.9 \\
 % & Raw Demonstrations      & 46.3 & \06.7 & 69.0 & 22.9 & 43.1 & 19.9 & 46.9 & 19.2 \\
 & DR \citep{liu2024toolnet}        & 51.6 & \02.6 & 60.0 & \09.7 & 41.6 & \05.7 & 45.1 & \06.8 \\
 & LR \citep{erdogan2024tinyagent}   & 50.5 & \07.1 & 56.6 &  20.3 & 42.8 & 17.3 & 42.1 & 17.3 \\
 & \cellcolor{verylightgray}\ourscl (Ours)  & \cellcolor{verylightgray}64.1 & \cellcolor{verylightgray}\cellcolor{verylightgray}11.8 & \cellcolor{verylightgray}60.2 & \cellcolor{verylightgray}26.4 & \cellcolor{verylightgray}\bfu{52.8} & \cellcolor{verylightgray}\bfu{21.3} & \cellcolor{verylightgray}\bfu{47.8} & \cellcolor{verylightgray}\bfu{22.5}\\
 & \cellcolor{verylightgray}\oursnn (Ours)       & \cellcolor{verylightgray}\bfu{78.4} &\cellcolor{verylightgray}\bfu{14.5} & \cellcolor{verylightgray}\bfu{83.1} & \cellcolor{verylightgray}\bfu{29.4} & \cellcolor{verylightgray}49.0 & \cellcolor{verylightgray}19.7 & \cellcolor{verylightgray}45.5 & \cellcolor{verylightgray}20.4 \\
\midrule
\multirow{5}{*}{\rotatebox[origin=c]{90}{Qwen3 4B~~}}
 & No ICL                                   & 59.6 & \05.0 & 83.5 & 13.5 & 45.8 & \05.1 & 54.3 & \06.8  \\
 % & Raw Demonstrations       & 60.2 & \03.6 & 87.6 & 26.2 & 49.8 & 10.7 & 57.3 & \09.8 \\
 & DR \citep{liu2024toolnet}       & 62.7 & \09.2 & 64.7 & 19.3 & 52.8 &  19.3 & 51.8 & 16.8  \\
 & LR \citep{erdogan2024tinyagent}   & 52.2 & \09.0 & 60.4 &  19.3 & 56.9 & 28.0 & 58.2 &  26.5 \\
 & \cellcolor{verylightgray}\ourscl (Ours)   & \cellcolor{verylightgray}68.1 & \cellcolor{verylightgray}14.4 & \cellcolor{verylightgray}64.7 & \cellcolor{verylightgray}27.9 & \cellcolor{verylightgray}56.0  & \cellcolor{verylightgray}\bfu{35.1} & \cellcolor{verylightgray}52.4 & \cellcolor{verylightgray}28.0 \\
 & \cellcolor{verylightgray}\oursnn (Ours)       & \cellcolor{verylightgray}\bfu{80.7} & \cellcolor{verylightgray}\bfu{16.9} & \cellcolor{verylightgray}\bfu{89.0} & \cellcolor{verylightgray}\bfu{31.8}  & \cellcolor{verylightgray}\bfu{60.5} & \cellcolor{verylightgray}29.8  & \cellcolor{verylightgray}\bfu{64.1} & \cellcolor{verylightgray}\bfu{30.9}  \\

\bottomrule
\end{tabular}
}

\caption{Comparison of selected methods in terms of function selection accuracy and end-to-end success rate over different datasets and models. All retrievers are paired with weighted Hard Masking for ICL. Best results per dataset and model are bolded. DTDR reaches higher accuracy as it retrieves a set of tools more closely aligned with $\funcF_t^*$.}
\label{table:aai_selection_and_calling}
\end{table*}

\paragraph{How to encode tools in prompt?} In Figure~\ref{fig:icl} we compare different ways to represent retrieved tools in the prompt. In three of four datasets, Hard and Soft Masking consistently outperform both the No ICL and the Raw Demonstration baselines. The exception is TaskBench DL, which lacks intrinsic function dependencies (see Table~\ref{table:datasets}), making dependency-based prompting less relevant.
% Soft Masking consistently matches or exceeds Raw Demonstrations, while 
Hard Masking achieves the highest accuracy for smaller models by directly pruning the set of available tools presented to the LLM.
Weighted Hard Masking performs comparably to its unweighted variant. Despite similar overall performance, we observe that weighted masking results in better accuracy when the retriever assigns high probability to the correct tools, and unweighted masking is better in the opposite scenarios. This suggests that unweighted masking should be preferred when expecting a large distribution shift between test queries and demonstration data, as predicted probabilities will be less accurate (details in Appendix~\ref{sec:app_weighted}). 
% From an efficiency standpoint, while Raw Demonstrations increase prompt length, Hard Masking reduces it substantially by removing descriptions of irrelevant tools. 

\begin{figure*}[!t]
    \centering
    \includegraphics[width=0.9\textwidth]{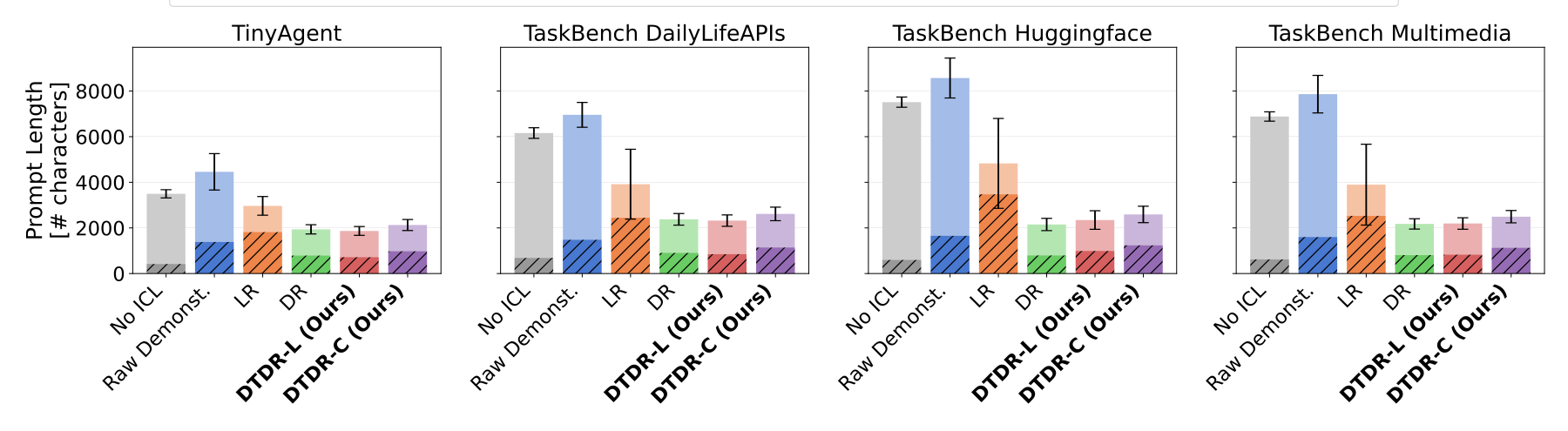}
    
    \caption{Prompt length across different methods and datasets. Our method reduces the prompt length by: 1) efficiently encoding ICL examples as tool dependencies instead of Raw Demonstrations, and 2) only retrieving tool dependencies which are relevant for the test query.}
    \label{fig:prompt}
   
\end{figure*}

\paragraph{How does DTDR perform on different datasets and model size?} 
In Table~\ref{table:aai_selection_and_calling}, we evaluate the Function Selection Accuracy (FSA) of best tool retrievers on several datasets and models of different sizes. Results on additional models and datasets can be found in Table \ref{table:allmodels2} in the Appendix. \ourscl consistently outperforms its static DR counterpart across all settings, while \oursnn is comparable or better than static LR. This supports our hypothesis that dynamic retrieval is key for high-quality function calling. Among all methods, \oursnn achieves the best overall performance. Notably, \oursnn applied to Qwen 3 4B and 8B respectively surpasses the No ICL baseline applied to Qwen 3 14B and GPT-4o, bridging the performance gap between edge and cloud models. In Figure \ref{fig:icl}, the Raw Demonstrations approach hurts smaller models but become competitive at larger scales, where LLMs exhibit stronger reasoning capabilities. However, including raw demonstrations in the prompt increases its length, which results in longer prefill time and worse compute efficiency. We analyze this in Figure~\ref{fig:prompt}, where we compare the prompt lengths obtained with different retrieval methods, distinguishing between constant and variable prompt sections. The constant section (guidelines, task demonstrations, and tool descriptions) can be precomputed and cached, while the variable section (query encoding and tool descriptions) must be recomputed for each query (see examples in Appendix~\ref{app:prompt}). Instead of Raw Demonstrations, we can directly use the retriever outputs to mask the set of tools provided to the LLM, which drastically reduces the prompt size. Static DR reduces the prompt length by selecting a subset of functions, but the subset changes at each step and must be updated dynamically in the prompt. Static LR can select a narrower set of tools, but it tends to overfit to retrieving the most frequent ones, which results in lower end-to-end performance. \ourscl and \oursnn can both reduce the size of the retrieved tool set and retain good downstream accuracy. In particular, they decrease the total prompt length by up to 73\% and the variable portion by up to 48\% when compared to raw text demonstrations. In terms of compute costs, \ours and most baseline retrievers use a small sentence encoder to represent the query, whose cost is negligible relative to the subsequent LLM prompting step. The only exception incurring overhead at inference time is the QTS retriever proposed by \citet{paramanayakam2025less}, which additionally prompts an LLM to describe the ideal tool set for the task.

\begin{figure*}[!h]
    \centering

    \includegraphics[width=0.9\textwidth]{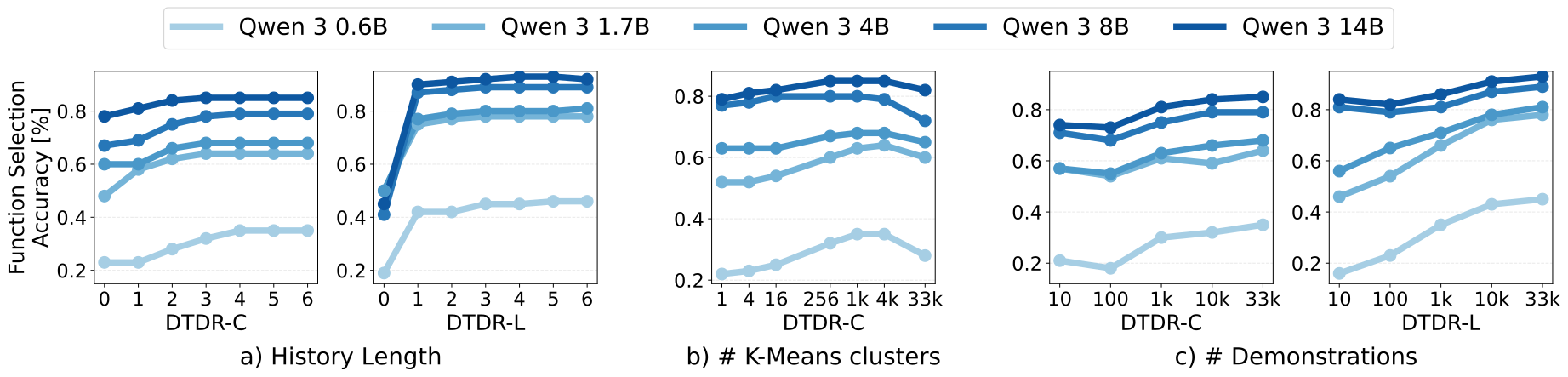}
    \caption{Ablations on: a) history length, b) \# of k-means clusters, and c) \# of demonstrations.}
    \label{fig:ablations}
 
\end{figure*}

\paragraph{Does DTDR improve end-to-end function calling success rates?} In Table~\ref{table:aai_selection_and_calling} we also report end-to-end Success Rate (SR) for the function calling task. We use the same model with different prompts for the function selection and parameter filling subtasks (details in Appendix~\ref{app:prompt}). If one or more mistakes occurs at any step in the plan (either in the function name or its parameters), the whole plan is evaluated as unsuccessful. For datasets with tool dependencies (TA, TB-HF, TB-MM), the dynamic variants of \ours improve the end-to-end Success Rate between 300\% and 600\% compared to the baseline without ICL, and between 15\% and 200\% compared to the best Tool Retrieval baseline.
Weaker baselines on smaller models fail on all datasets. LR \citep{erdogan2024tinyagent} tends to overfit to predicting the most common function in the dataset, while the same architecture in its dynamic variant \oursnn (adding function call history as input), prevents this problem and yields significantly better performance. 
% On the other hand, the clustering-based variant \ourscl is based on the statistics of demonstration data, and does not have the issue of overfitting to most common functions. 
Using Raw Demonstrations yields good performance with larger models, as they have increased capabilities for longer context. However, as previously discussed, this approach would yield significant increases in prefill latency.

\paragraph{Impact of History Length, Clustering, and Number of Demonstrations.} Figure~\ref{fig:ablations}a examines the effect of varying the history length $l$ in the function call sequence $[f_{t-l}, ..., f_{t-1}]$. Longer histories enhance dependency modeling in \ourscl and provide richer context for \oursnn. Both models benefit up to $l=3$, beyond which gains taper off. Without history, \oursnn tends to overfit, often defaulting to the most frequent functions, as previously observed. Figure~\ref{fig:ablations}b explores query-conditioning by varying the number of clusters in \ourscl. With a single cluster, retrieval is query-agnostic, similar to \citep{liu2024toolnet}. Increasing the number of clusters enables more targeted retrieval from demonstrations similar to the test query. The extreme case (nearest-neighbor retrieval) assigns one demonstration per cluster. We find optimal performance when the number of clusters is approximately $1/10$ of the demonstrations, which we adopt as the default setting for all \ourscl experiments. Figure~\ref{fig:ablations}c shows how increasing the number of demonstrations influences Function Selection Accuracy for both \ourscl and \oursnn. Performance improves with more data, plateauing around 10k samples. Gains are more pronounced for smaller LLM backbones. Notably, \oursnn suffers a sharp drop below 1k samples due to overfitting, highlighting its sensitivity compared to the more robust \ourscl.

\paragraph{How does accuracy scale with increasing Plan Length?} We observe that in Table \ref{tab:plan_length_analysis}, DR remains relatively stable across plan lengths, as it conditions only on the most recent tool call. LR performs well on short plans but degrades as plan length increases. DTDR-C outperforms both baselines on shorter plans (length < 6) but declines on longer plans, likely due to limited data coverage for higher-order dependencies. DTDR-L consistently achieves the highest accuracy across lengths and generalizes better to longer plans, suggesting stronger extrapolation of dependency structure.

\begin{table}[t]
\centering
\small
\resizebox{\columnwidth}{!}{\begin{tabular}{l|cccccccc}
\toprule
& \multicolumn{8}{|c}{\textbf{Plan Length}} \\
\midrule 
\textbf{Method} & \textbf{2} & \textbf{3} & \textbf{4} & \textbf{5} & \textbf{6} & \textbf{7} & \textbf{8} & \textbf{9} \\
\midrule
DR      & 44.6 & 36.5 & 45.1 & 53.7 & 62.7 & 61.6 & 51.7 & 50.6 \\
LR      & 50.0 & 34.3 & 26.1 & 21.7 & 19.0 & 16.6 & 15.6 & 12.9 \\
DTDR-C  & 88.6 & 67.7 & 62.6 & 60.3 & 63.4 & 54.6 & 48.8 & 49.7 \\
DTDR-L  & 96.3 & 88.1 & 87.6 & 88.5 & 88.5 & 87.2 & 84.8 & 85.1 \\
\midrule
\# Samples & 1041 & 988 & 995 & 1032 & 988 & 564 & 263 & 105 \\
\bottomrule
\end{tabular}}
\caption{Top-1 tool retrieval accuracy (\%) as a function of plan length in TinyAgent.}
\label{tab:plan_length_analysis}
\end{table}

\section{Conclusion}
We introduce Dynamic Tool Dependency Retrieval (DTDR), a retrieval method that leverages both the tool dependencies and historical predictions via a retrieval module trained on demonstrations. DTDR has two variants: a supervised NN-based approach and an unsupervised clustering-based approach, both suitable for low-resource, on-device settings. Experiments across multiple datasets and LLM backbones show that DTDR significantly improves retrieval quality and downstream function selection over baselines.  

\section{Limitations}

We wish to highlight some failure modes of DTDR:

\begin{itemize}
    \item \textbf{Out-of-distribution generalization (DTDR-C).} DTDR-C performance degrades on out-of-distribution tasks, as both the clustering mechanism and higher-order Markov modeling require sufficient training data coverage to reliably estimate transition structure. Limited coverage reduces cluster quality and transition reliability.
    \item \textbf{Niche or non-standard tool vocabularies.} DTDR-C performance decreases on datasets with uncommon or highly domain-specific tool names and descriptions (e.g., HuggingFace-style function names). In such cases, pre-trained embeddings used for clustering may not adequately capture semantic similarity. A learnable model such as DTDR-L mitigates this issue by adapting its weights to better align with task-specific embedding representations.
    \item \textbf{Strong OOD tool dependencies.} All retrieval-based methods fail under strong out-of-distribution conditions, where tool names or dependency structures are entirely unseen during training. In such cases, an uncertainty-aware fallback mechanism could be applied: when the retriever’s confidence is low, all tools are passed to the LLM instead of filtering. For DTDR-C, uncertainty can be estimated via cluster assignment likelihood; for DTDR-L, feature-space distances such as the Mahalanobis distance [1] can be used for OOD detection.
\end{itemize}

Additional failure modes can be found in Appendix \ref{sec:app_add_results} and \ref{sec:app_weighted}. Furthermore, this work assumes access to expert, ground-truth demonstrations. While demonstrations may be available as previously mentioned, having them be totally correct for all tasks may be a restrictive requirement. Therefore, future work can try to learn from failed, incorrect, or unlabeled samples. Other future directions include extending to multimodal tool-based tasks (e.g. robotics) and adapting to evolving tool sets.

\bibliography{custom}

\appendix

\begin{table*}[t]
\centering
\begin{tabular}{l|cccc}
\toprule
 \multirowcell{2}{Dataset} & \multirowcell{2}{\# Plans \\} & \multirowcell{2}{\# Tools \\ } & \multirowcell{2}{\# Tool calls\\per plan} & \multirowcell{2}{\# Tool dependencies\\per plan}\\
 & \\
\midrule
TinyAgent & 39874 & 17 & 4.5\spm1.9 & 1.9\spm1.8 \\
TaskBench DailyLife & 3860 & 41 & 4.1\spm1.7 & 0.1\spm0.2 \\
TaskBench HuggingFace & 4959 & 24 & 3.2\spm1.5 & 1.1\spm1.5 \\
TaskBench Multimedia & 4310 & 41 & 3.6\spm1.7 &  1.5\spm1.7\\
\bottomrule
\end{tabular}

\caption{Function Calling Benchmark statistics. The number of tool dependencies refers to cases where the parameter of a function call is the output of a function call earlier in the plan. Notice that TaskBench DailyLife mainly contains plans with unrelated function calls.}
\label{table:datasets}
\end{table*}

\section{Detailed Related Works} \label{sec:app_related_works}

\paragraph{Training LLMs for Tool Use.}
Recent LLMs demonstrate impressive tool-usage capabilities, with cloud-based models such as GPT-5 \citep{openai2025gpt5systemcard}, Claude 4 \citep{anthropic2025claude4systemcard}, and GLM-4.5 \citep{zeng2025glm} leading by a wide margin on function-call benchmarks like BFCL V4 \citep{patilberkeley} and $\tau^2$-Bench \citep{barres2025tau}. In contrast, smaller LLMs designed for edge deployment perform significantly worse on agentic tasks when evaluated out-of-the-box, creating a challenge for widespread adoption on mobile devices. To address this, recent work has explored fine-tuning LLMs on large function calling datasets, sourced from cloud providers \citep{gorilla-openfunctions-v2}, generated by cloud models \cite{qiao2023making}, or obtained through self-play in agentic environments \citep{schick2023toolformer}. However, collecting tool-use data for fine-tuning is often prohibitively expensive, and these fine-tuned models frequently struggle to generalize to new or updated tools.

\paragraph{LLM Prompting with Tool Retrieval.}
An alternative line of research leverages the in-context learning capabilities of LLMs, introducing prompt-engineering strategies for tool learning. This can involve including tool descriptions \citep{shen2023hugginggpt, shen2024taskbench} or example trajectories \citep{paranjape2023art,sarukkai2025self} within the model prompt. To handle hundreds of apps and APIs, recent work uses retrieval-augmented generation to sub-select tools that are relevant to the task. ToolLLM \citep{qin2023toolllm} and ToolReAGt \citep{braunschweiler2025toolreagt} identify the most suitable tools by comparing embeddings derived from the user query and tool documentation. Less-is-More \citep{paramanayakam2025less} further prompts an LLM to generate documentation for the ideal tools to solve the query. ART \citep{paranjape2023art} uses LLM prompting to retrieve example trajectories for most relevant tools.

\paragraph{Tool Dependency Retrieval.}
The execution of certain tools often depends on the outputs of others. ControlLLM \citep{liu2024controlllm} and SciToolAgent \citep{ding2025scitoolagent} address specific scenarios where the tool dependency graph is known. %, prompting an LLM to plan over this graph to complete the task. 
In most real-world cases, however, the set of tools and their dependencies are not known in advance or may change at test time, for example when apps are installed or updated. In that case, tool dependencies could be learned from demonstration data provided by tool vendors \citep{paranjape2023art}, extracted from historical tool trajectories \citep{chen2025gtool}, generated procedurally \citep{erdogan2024tinyagent, shen2024taskbench}, or synthesized using cloud LLMs \citep{gao2025tool, patil2024gorilla,qin2023toolllm}. Tool Graph Retriever \citep{gao2025tool} and GTool \citep{chen2025gtool} employ GNNs to encode the learned dependencies, using the resulting embeddings for similarity-based retrieval and for LLM prompting. In contrast, ToolNet \citep{liu2024toolnet} parses the tool dependency graph to identify the set of eligible tools after the latest tool call in the plan. Rather than modeling a static tool graph from demonstration data, we propose a lightweight approach that dynamically retrieves specific tool dependencies which are relevant to the test query and the sequence of previously called tools.

\section{Dataset Statistics}\label{app:datasets}
In Table~\ref{table:datasets} we report important statistics for the datasets considered in our evaluation. While the average length of the plans is comparable across datasets, the number of tool dependencies varies significantly, which means that certain datasets benefit more than others in modeling and employing tool dependencies for function selection.

\begin{table*}[h]
    \centering
    \resizebox{\textwidth}{!}{\begin{tabular}{c|c|c}
        \textbf{Tool Dependency Graph } & \textbf{Unweighted Tool},  & \textbf{Weighted Tool} \\ 
        \textbf{ $G_6$ Key, $f_6$} & \textbf{ Dependency Graph Value}, $F^{f_6}$  & \textbf{Dependency Graph Value,} $P(f'|f_6)$ \\ \hline
        (`start', & [`join', & \{`join : $0.8$, \\
       `start',  &  `create\_calendar\_event',  & `create\_calendar\_event' : $0.05$, \\
        `start', & `create\_reminder', &  `create\_reminder' : $0.05$, \\
        `open\_and\_get\_file\_path', & `summarize\_pdf', & `summarize\_pdf': $0.05$,\\
        `get\_email\_address', & `append\_note\_content'] & `append\_note\_content': $0.05$ \}\\
        `compose\_new\_email') &  & \\
    \end{tabular}}
    \caption{\textbf{Example Tool Dependency Graph.} Given a function sequence of length $6$, $f_6$ (left), we can compute based on demonstration data, $D$, the unweighted tool dependency graph (middle) and weighted tool dependency graph (right).}
    \label{tab:tool_dependency_graph_example}
\end{table*}

\section{Implementation Details} \label{app:implementation}

Please see Appendix ~\ref{app:pseudocode} for pseudocode to implement these methods.

\paragraph{Tool Dependency Graph Construction in \ourscl}
We compute the tool dependency graph based on the training data $D_{train}$. Each training datapoint contains a ground-truth trajectory for the given query. Let $f_l = (f_{i-o}, f_{i-l+1}, \cdots, f_i)$ be a history of function calls of length $l$. For example, a possible $f_2$ could be the tuple, ('get\_email', 'reply\_email'), where `get\_email' is the first function in the sequence and `reply\_email is the latest function. For each function sequence $f_l$ used in the ground-truth trajectories of $D_{train}$, let $\funcF^{f_l}$ be the set of all unique functions that have appeared immediately after $f_l$.  The tool dependency graph $G_l$ is formatted according to whether it is probabilistic or not. When $G_l$ is probabilistic, each $f_l$ is a key in a nested dictionary. The value of each $f_l$ is another dictionary representing a probability distribution where the inner keys are all the functions in $\funcF^{f_l}$. For each inner function key, $f' \in \funcF^{f_l}$, the value is, $P(f'| f_l)$: the percent of times $f'$ appears given that $f_l$ appears immediately before it. Thus, $P(f'|f_l)$ can be considered an $l$-ordered Markov Chain. If $G_l$ is unweighted, then $G_l$ is a flat dictionary where each $f_l$ is a key and the corresponding value is $\funcF^{f_l}$. Table \ref{tab:tool_dependency_graph_example} gives and example of a tool dependency graph of order $6$, or $G_6$.  To pad function sequences with length less than order $l$, we use the string `start' as a placeholder. In Table \ref{tab:tool_dependency_graph_example}, the example $f_6$ only has three functions in the history, so we use use three `start' strings in the beginning. In summary, we use a sliding window approach with window length $l$, and we construct $G$ based on the next function right after the current window. For \emph{\ourscl}, we find the optimal number of clusters to be $1/10$ of the training set size.

\paragraph{Classifier optimization in \oursnn}  Fine-tuning for TinyAgent and \emph{\oursnn} variants is performed using Adam with a learning rate of $10^{-3}$, learning rate decay of $0.9$, weight decay of $10^{-5}$, and $10$ epochs. For \oursnn, the training of $\phi$ is per-step. For each training query $q \in Q_{train}$, we train our predictor $\phi$ using the sum of independent $|\funcF|$ binary cross-entropy loss terms. Formally, let $d_{0:t-1}$ be the possible ground-truth demonstration trajectory for $q$ up until time $t$. Both $d_{0:t-1}$ and $f_{0:t-1}$ are a list of function names, so $d_{0:t-1}$ can be concatenated to $q$ and passed into $\phi$. Therefore, we can write

\small
\begin{equation}
\begin{split}\label{eq:bce_loss}
&\phi^*=  \\& \argmin_{\phi}-\sum_{q \in Q_{train}} \sum_{t=0}^T \sum_{f \in \funcF} (\mathbf{1}_{f \in \funcF_{q,t}^*})(\log \phi(q, d_{0:t-1})).
\end{split}
\end{equation}
\normalsize

\begin{table*}[!h]
\centering
\resizebox{0.8\textwidth}{!}{%
\begin{tabular}{c|l|c|cc}
\toprule
&  & \multirowcell{3}{\textbf{Function}\\\textbf{Selection}\\\textbf{Acc. [\%] ($\uparrow$)}} & & \\
 & & & \multicolumn{2}{c}{\textbf{Retrieval Metrics}} \\
& \textbf{Retrieval Method} &   & ~~\textbf{MRR} \textbf{($\uparrow$)}~~ & ~~\textbf{F$_1$} \textbf{($\uparrow$)}~~ \\ % & \textbf{FLOPS ($\downarrow$)} \\
\midrule
\multirow{5}{*}{\rotatebox[origin=c]{90}{\textbf{TinyAgent}~~~~~~~}}
 & Random Guess & \05.9& 0.26& 0.12 \\
 & BM-25 \citep{robertson2009probabilistic} & 23.1& 0.35& 0.20 \\
 & QTS (Vanilla) & 15.8&	0.26 &	0.14 \\
 & QTS \citep{gao2025tool}  & 21.5&	0.18 &	0.09 \\
 & QTS \citep{paramanayakam2025less} & 23.7&	0.36&	0.19 \\
 & DR \citep{liu2024toolnet} & 30.7&	0.70& 0.49 \\
 & \cellcolor{verylightgray}Dynamic DR (\ourscl) (Ours) & \cellcolor{verylightgray}43.3&	\cellcolor{verylightgray}0.78 &	\cellcolor{verylightgray}\bfu{0.56}\\
 & LR \citep{erdogan2024tinyagent} & 25.6	 &0.53&	0.39\\
 & \cellcolor{verylightgray}Dynamic LR (\oursnn) (Ours)  & \cellcolor{verylightgray}\bfu{65.1}&	\cellcolor{verylightgray}\bfu{0.93} &	\cellcolor{verylightgray}0.55\\
 \midrule
\multirow{5}{*}{\rotatebox[origin=c]{90}{\textbf{Taskbench-HF}~~~~}}
 & Random Guess & \04.2& 0.16& 0.05\\
 & BM-25 \citep{robertson2009probabilistic} & \08.3	&0.18	& 0.05\\
 & QTS (Vanilla)   & \07.9	&0.21&	0.08\\
 & QTS \citep{gao2025tool} & 14.6&	0.37	& 0.27\\
 & QTS \citep{paramanayakam2025less}  & 12.9 &	0.30	& 0.18\\
 & DR \citep{liu2024toolnet}  & 17.1 & 0.46& 0.33\\
 & \cellcolor{verylightgray}Dynamic DR (\ourscl) (Ours)  &  \cellcolor{verylightgray}27.5	&\cellcolor{verylightgray}0.64	&\cellcolor{verylightgray}0.55\\
 & LR \citep{erdogan2024tinyagent}  & 20.5	&0.53 & 0.32\\
 & \cellcolor{verylightgray}Dynamic LR (\oursnn) (Ours)  &  \cellcolor{verylightgray}\bfu{27.8}&	\cellcolor{verylightgray}\bfu{0.75}	&\cellcolor{verylightgray}\bfu{0.63}\\
 \midrule
\multirow{5}{*}{\rotatebox[origin=c]{90}{\textbf{Taskbench-MM}~~~}}
 & Random Guess & \02.4 & 0.11& 0.03\\
 & BM-25 \citep{robertson2009probabilistic} &  13.7	&0.26	&	0.19\\
 & QTS (Vanilla)    & \05.3	&0.14	&0.04\\
 & QTS \citep{gao2025tool} &  \05.3	&0.14	&0.20\\
 & QTS \citep{paramanayakam2025less}  & \07.6 &	0.29	&	0.18\\
 & DR \citep{liu2024toolnet}  & 13.3	&0.38	&0.27	\\
 & \cellcolor{verylightgray}Dynamic DR (\ourscl) (Ours)  &  \cellcolor{verylightgray}24.4	&\cellcolor{verylightgray}0.52	&\cellcolor{verylightgray}0.43\\
 & LR \citep{erdogan2024tinyagent}  &  21.0	&0.49 & 0.28\\
 & \cellcolor{verylightgray}Dynamic LR (\oursnn) (Ours)  &  \cellcolor{verylightgray}\bfu{27.0}&	\cellcolor{verylightgray}\bfu{0.69}	&\cellcolor{verylightgray}\bfu{0.55}\\
 \midrule
\multirow{5}{*}{\rotatebox[origin=c]{90}{\textbf{Taskbench-DL}~~~~}}
 & Random Guess & \02.4& 0.15& 0.07 \\
 & BM-25 \citep{robertson2009probabilistic} & 13.5&0.31&	0.15\\
 & QTS (Vanilla)  & \07.5&	0.16 &	0.07	\\
 & QTS \citep{gao2025tool}  & 13.3&		0.17 &	0.06\\
 & QTS \citep{paramanayakam2025less} & 18.9	&0.40&	0.26\\
 & DR \citep{liu2024toolnet} & 14.7 & 0.36&	0.18\\
 & \cellcolor{verylightgray}Dynamic DR (\ourscl) (Ours) & \cellcolor{verylightgray}28.2&\cellcolor{verylightgray}0.54	&\cellcolor{verylightgray}0.29 \\
 & LR \citep{erdogan2024tinyagent} & 23.6	&	0.55 & 0.44\\
 & \cellcolor{verylightgray}Dynamic LR (\oursnn) (Ours)  & \cellcolor{verylightgray}\bfu{58.9}&	\cellcolor{verylightgray}\bfu{0.85} &	\cellcolor{verylightgray}\bfu{0.64}\\
\bottomrule
\end{tabular}
}
\caption{Retrieval performance for methods from different categories (QTS = Query-Tool Similarity, LR = Learned Retriever, DR = Dependency Retriever). QTS and LR categories are query-conditioned, while DR is history-conditioned. Our Dynamic methods are conditioned on both query and tool history. Function Selection Accuracy is reported on Qwen 3 0.6B using hard masking as ICL method. Best results per dataset and metrics are in bold.}
\label{table:retrieval_Amir_appendix}
\end{table*}

\begin{table*}[!h]
\centering
\resizebox{0.9\textwidth}{!}{%
\begin{tabular}{l|l|cccc}
\toprule
 \multicolumn{6}{c}{\textbf{Function Selection Accuracy [\%] ($\uparrow$)}}\\
\midrule
    % & \textbf{ICL Method} & \multicolumn{4}{c}{\textbf{Function Selection Accuracy ($\uparrow$)}} \\
 & \multirowcell{2}{\textbf{Method}} & \multirowcell{2}{TinyAgent} & \multirowcell{2}{TaskBench \\ DailyLife} & \multirowcell{2}{TaskBench \\ HuggingFace} & \multirowcell{2}{TaskBench \\ Multimedia} \\
 & \\
\midrule
\multirow{5}{*}{\rotatebox[origin=c]{90}{Qwen3 0.6B~~}}
 & No ICL                                   & 23.2 & 12.8 & \05.5 & \07.3 \\
 & Dependency Retrieval \citep{liu2024toolnet}       & 22.0 & 11.5 & \06.7 & \09.5 \\
 & Learned Retrieval \citep{erdogan2024tinyagent}   & 19.8 & 19.6 & 10.0 & 12.1 \\
 & \cellcolor{verylightgray}\ourscl (Ours)  & \cellcolor{verylightgray}35.3 & \cellcolor{verylightgray}23.7 & \cellcolor{verylightgray}\bfu{20.8} & \cellcolor{verylightgray}\bfu{20.2} \\
 &  \cellcolor{verylightgray}DTDR-C: Raw Dems Prompting (Ours)  & \cellcolor{verylightgray}17.2 & \cellcolor{verylightgray}10.5 & \cellcolor{verylightgray}\04.3 & \cellcolor{verylightgray}\05.3 \\
 & \cellcolor{verylightgray}\oursnn (Ours)       & \cellcolor{verylightgray}\bfu{46.1} & \cellcolor{verylightgray}\bfu{45.9} & \cellcolor{verylightgray}18.9 & \cellcolor{verylightgray}17.6 \\
\midrule
\multirow{5}{*}{\rotatebox[origin=c]{90}{Qwen3 1.7B~~}}
 & No ICL                                   & 48.1 & 66.7 & 31.2 & 38.1 \\
 & Dependency Retrieval \citep{liu2024toolnet}        & 51.6 & 60.0 & 41.6 & 45.1 \\
 & Learned Retrieval \citep{erdogan2024tinyagent}   & 50.5 & 56.6 & 42.8 & 42.1 \\
 & \cellcolor{verylightgray}\ourscl (Ours)  & \cellcolor{verylightgray}64.1 & \cellcolor{verylightgray}60.2 & \cellcolor{verylightgray}\bfu{52.8} & \cellcolor{verylightgray}\bfu{47.8} \\
  & \cellcolor{verylightgray}DTDR-C: Raw Dems Prompting (Ours) & \cellcolor{verylightgray}46.3 & \cellcolor{verylightgray}69.0 & \cellcolor{verylightgray}43.1 & \cellcolor{verylightgray}46.9 \\
 & \cellcolor{verylightgray}\oursnn (Ours)       & \cellcolor{verylightgray}\bfu{78.4} & \cellcolor{verylightgray}\bfu{83.1} & \cellcolor{verylightgray}49.0 & \cellcolor{verylightgray}45.5 \\
\midrule
\multirow{5}{*}{\rotatebox[origin=c]{90}{Qwen3 4B~~}}
 & No ICL                                   & 59.6 & 83.5 & 45.8 & 54.3 \\
 & Dependency Retrieval \citep{liu2024toolnet}       & 62.7 & 64.7 & 52.8 & 51.8 \\
 & Learned Retrieval \citep{erdogan2024tinyagent}   & 52.2 & 60.4 & 56.9 & 58.2 \\
 & \cellcolor{verylightgray}\ourscl (Ours)   & \cellcolor{verylightgray}68.1 & \cellcolor{verylightgray}64.7 & \cellcolor{verylightgray}56.0 & \cellcolor{verylightgray}52.4 \\
 & \cellcolor{verylightgray}DTDR-C: Raw Dems Prompting (Ours)    & \cellcolor{verylightgray}60.2 & \cellcolor{verylightgray}87.6 & \cellcolor{verylightgray}49.8 & \cellcolor{verylightgray}57.3 \\
 & \cellcolor{verylightgray}\oursnn (Ours)       & \cellcolor{verylightgray}\bfu{80.7} & \cellcolor{verylightgray}\bfu{89.0} & \cellcolor{verylightgray}\bfu{60.5} & \cellcolor{verylightgray}\bfu{64.1} \\
\midrule
\multirow{5}{*}{\rotatebox[origin=c]{90}{Qwen3 8B~~}}
& No ICL                                    & 71.3 & 93.9 & 70.5 & 78.5 \\

 & Dependency Retrieval \citep{liu2024toolnet}        & 76.7 & 75.3 & 67.3 & 69.1 \\
 & Learned Retrieval \citep{erdogan2024tinyagent}   & 43.7 & 52.2 & 52.0 & 49.6 \\
 & \cellcolor{verylightgray}\ourscl (Ours)  & \cellcolor{verylightgray}80.4 & \cellcolor{verylightgray}75.3 & \cellcolor{verylightgray}67.3 & \cellcolor{verylightgray}69.1 \\
  & \cellcolor{verylightgray}DTDR-C: Raw Dems Prompting (Ours)      & \cellcolor{verylightgray}67.8 & \cellcolor{verylightgray}\bfu{94.5} & \cellcolor{verylightgray}\bfu{72.5} & \cellcolor{verylightgray}\bfu{80.4} \\
 & \cellcolor{verylightgray}\oursnn (Ours)       & \cellcolor{verylightgray}\bfu{89.0} & \cellcolor{verylightgray}91.0 & \cellcolor{verylightgray}63.8 & \cellcolor{verylightgray}67.6 \\
\midrule
\multirow{5}{*}{\rotatebox[origin=c]{90}{Qwen3 14B~~}}
 & No ICL                                   & 78.0 & 96.2 & 74.4 & 81.7 \\
 & Dependency Retrieval \citep{liu2024toolnet}        & 79.5 & 84.1 & 69.2 & 73.8 \\
 & Learned Retrieval \citep{erdogan2024tinyagent}   & 51.1 & 56.5 & 53.5 & 50.8 \\
 & \cellcolor{verylightgray}\ourscl (Ours)   & \cellcolor{verylightgray}84.8 & \cellcolor{verylightgray}84.1 & \cellcolor{verylightgray}71.5 & \cellcolor{verylightgray}74.0 \\
 & \cellcolor{verylightgray}DTDR-C: Raw Dems Prompting (Ours) & \cellcolor{verylightgray}77.9 & \cellcolor{verylightgray}\bfu{96.6} & \cellcolor{verylightgray}\bfu{76.7} & \cellcolor{verylightgray}\bfu{82.3} \\
 & \cellcolor{verylightgray}\oursnn (Ours)       & \cellcolor{verylightgray}\bfu{92.5} & \cellcolor{verylightgray}93.7 & \cellcolor{verylightgray}71.2 & \cellcolor{verylightgray}73.4 \\
\midrule
\multirow{5}{*}{\rotatebox[origin=c]{90}{Gorilla V2~~}}
 & No ICL                                   & 39.5 & PLE & PLE & PLE \\

 & Dependency Retrieval \citep{liu2024toolnet}         & 32.2 & PLE & PLE & PLE \\
 & Learned Retrieval \citep{erdogan2024tinyagent}   & 37.9 & PLE & PLE & PLE \\
 & \cellcolor{verylightgray}\ourscl (Ours)   & \cellcolor{verylightgray}43.4 & \cellcolor{verylightgray}PLE & \cellcolor{verylightgray}PLE & \cellcolor{verylightgray}PLE \\
 & \cellcolor{verylightgray}DTDR-C: Raw Dems Prompting (Ours) & \cellcolor{verylightgray}PLE & \cellcolor{verylightgray}PLE & \cellcolor{verylightgray}PLE & \cellcolor{verylightgray}PLE \\
 & \cellcolor{verylightgray}\oursnn (Ours)       & \cellcolor{verylightgray}\bfu{58.7} & \cellcolor{verylightgray}PLE & \cellcolor{verylightgray}PLE & \cellcolor{verylightgray}PLE \\
 \midrule
\multirow{5}{*}{\rotatebox[origin=c]{90}{GPT-4o~~}}
 & No ICL                                   & 84.7 & 96.6 & 74.8 & 81.6 \\

 & Dependency Retrieval \citep{liu2024toolnet}         & 85.9 & 82.9 & 70.0 & 72.8 \\
 & Learned Retrieval \citep{erdogan2024tinyagent}   & 53.2 & 48.1 & 55.2 & 51.1 \\
 & \cellcolor{verylightgray}\ourscl (Ours)   & \cellcolor{verylightgray}85.9 & \cellcolor{verylightgray}82.9 & \cellcolor{verylightgray}71.9 & \cellcolor{verylightgray}74.8 \\
 & \cellcolor{verylightgray}DTDR-C: Raw Dems Prompting (Ours)  & \cellcolor{verylightgray}86.8 & \cellcolor{verylightgray}\bfu{96.8} & \cellcolor{verylightgray}\bfu{77.1} & \cellcolor{verylightgray}\bfu{82.3} \\
 & \cellcolor{verylightgray}\oursnn (Ours)   & \cellcolor{verylightgray}\bfu{94.2} & \cellcolor{verylightgray}91.5 & \cellcolor{verylightgray}71.8 & \cellcolor{verylightgray}73.7 \\
\bottomrule
\end{tabular}
}
\caption{Comparison of selected methods in terms of function selection accuracy over different datasets and models. Best results per dataset and model are bolded. Gorilla V2 has 4096 maximum context length, which easily results in PLE (Prompt Length Exceeded).}
\label{table:allmodels2}

\end{table*}

\paragraph{Other details and hyper-parameters}For all Tool Retriever methods, we use the same sentence-embedding model \emph{paraphrase-MiniLM-L6-v2} \citep{reimers2019sentencebert} across all approaches to ensure comparability. The dimensionality of the embeddings $e_\theta = 384$. For \emph{Less-is-More}, we ensure the same LLM is used both for Function Selection and for recommending the ideal tool descriptions. We consider Search Level 1 but not Level 2, as the latter needs interaction with a cloud model, which is prohibitive on-device. For \emph{ToolNet}, we build the tool dependency graph from the training split and use it to retrieve and score relevant tools based on the latest tool call in the plan. For a fair comparison with \emph{Tool Graph Retriever}, we also use the available example trajectories to build the tool dependency graph, instead of relying on a separate predictor. For the \emph{TinyAgent} retriever, we fine-tune a linear layer on top of the sentence embedding model to score all tools given the test query. For all Tool Retrievers other than \emph{\ourscl} (which does not require it) we sweep the function selection threshold with values of $0.2, 0.23, 0.5$ and set it based on function selection accuracy. For ICL experiments we use greedy decoding with a maximum generation length matching the longest tool name in the dataset. For the \emph{Raw Demonstrations} ICL method, we provide 5 random demonstrations as examples. When defining train and test splits for each dataset, we ensure that the number of occurrences in train and test data for each tool follows then 70/30 ratio. We use FP16 precision for Qwen models and Gorilla-V2, while GPT-4o is prompted through cloud APIs. All experiments are conducted on NVIDIA A100 GPUs.

\section{Metric Details}\label{sec:metric_details}

For downstream performance, we measure \emph{Function Selection Accuracy} as described in \citep{erdogan2024tinyagent}: for each step, the selected function is considered correct only it belongs to the set of acceptable tools according to the Directed Acyclic Graph (DAG) of the ground-truth plan. For \emph{Success Rate} we evaluate both function selection \emph{and} predicted parameters for \emph{all} steps in the plan. The success rate is 1 only if the DAG of the plan is isomorphic to the DAG of the ground-truth plan \citep{erdogan2024tinyagent}. The same LLM backbone is prompted to predict the function calling parameters given the query and selected function. Example of prompts for function selection and parameter filling are provided in Appendix \ref{app:prompt}.
Tool Retrieval methods are evaluated with ranking and retrieval metrics. \emph{Mean Reciprocal Rank (MRR)} measures the relative ranking of tools. $\mathit{F_1}$ score is the harmonic mean of Precision and Recall and is computed considering the top-$k$ highest-scoring tools as retrieved, with $k$ being the number of acceptable tools according to the DAG of the ground-truth plan.

\section{Additional Results}\label{sec:app_add_results}
\label{app:full_table}

\paragraph{Retrieval Accuracy} We report the full values for retrieval and function selection accuracy with Qwen 3 0.6B with unweighted Hard Masking. \ours consistently outperforms in both retireval and function selection across datasets. We note that for Taskbench-Multimedia that LR outperforms in function selection due to overfitting to repeatedly predict ``end-of-plan'' which is always the last function in each task.

\paragraph{Function Selection Accuracy Across Models}
We report here the full results on function selection accuracy discussed in the Results and Discussion Section. Additional LLM backbones include Qwen 3 14B and Gorilla-v2. At similar footprints, Gorilla V2 under-performs Qwen 3 8B, likely due to differences between our test data and the coding API data used for Gorilla’s fine-tuning. This difference in performance reinforces our claim that adaptive ICL strategies are preferable to fine-tuned generalist models, as ICL can seamlessly adapt to unseen tools at test time.

\section{Weighted vs Unweighted Masking}\label{sec:app_weighted}
To see the impact between weighted and unweighted masking, we compute the percentage of correct predictions when the $\ourscl$-based $\omega$ did not retrieve any $f \in F^*_t$ with a probability above threshold $0.5$. We see for Tinyagent across all models and number of cluster settings, Hard Masking unweighted achieves the maximum of around $33.6\%$ and outperforms hard mask weighted by up to $10\%$. This difference is also present when comparing Soft Masking with its weighted variant, and this difference shows that when the demonstration data does not reflect the task well enough, the weights provided in the prompt mislead the agent. However, as the number of clusters increase, this difference between in values for weighted and unweighted decreases. Similar observations are present for probability thresholds of $0.4$ and $0.3$. 

\section{Prompt Examples} \label{app:prompt}

We provide example prompts showing guidelines and ICL techniques used for function selection and parameter filling. Grayed out comments are only used as reference, and are not included in the actual prompt.

\begin{promptbox}{Function Selection}
(*@\textit{\textcolor{gray}{\# System Guidelines}}@*)
You are an expert AI Agent specialized in creating plans comprised of function calls to
satisfy user queries.
You are given a user query and a set of functions that should be used to solve the query.
You are also given the executed plan so far to solve the query.
You task is to only select one of the given functions as the immediate next step to be added to the plan.
You MUST only output one of the functions among the function list.
You MUST NOT include any other text in the response.
You are given below an example query along with the executed plan paired with the expected answer to solve the query.
You can use this example to learn to predict the correct function to add to the plan for the user query.

(*@\textit{\textcolor{gray}{\# Example Functions}}@*)
Set of available functions and their descriptions:
[{'function': 'create_calendar_event', 'description': 'Create a calendar event'}, 
..., 
{'function': 'summarize_pdf', 'description': 'Summarizes the content of a PDF file and returns the summary'}]
You will choose one of the function names above.

(*@\textit{\textcolor{gray}{\# Example Task}}@*)
User query: 'Open the file "ProjectPlan.docx", create a reminder to review it by next Monday, and find directions to the nearest cafe in Apple Maps.'.
Executed plan so far:
out_1 = open_and_get_file_path(name_or_path='ProjectPlan.docx')

(*@\textit{\textcolor{gray}{\# Example Answer}}@*)
create_reminder

(*@\textit{\textcolor{gray}{\# Test Functions}}@*)
Set of available functions and their descriptions:
[{'function': 'create_calendar_event', 'description': 'Create a calendar event'}, 
..., 
{'function': 'summarize_pdf', 'description': 'Summarizes the content of a PDF file and returns the summary'}]
You will choose one of the function names above.

(*@\textit{\textcolor{gray}{\# Test Task}}@*)
Now choose the next function call for this task:
User query: 'Reply to the currently selected email in Mail with the match details attached and create a new note titled "Festival Notes" to summarize the discussions.'.
Executed plan so far: None
\end{promptbox}

\begin{promptbox}{Parameter Filling}
(*@\textit{\textcolor{gray}{\# System Guidelines}}@*)
You are an expert AI Agent specialized in calling functions to satisfy user queries.
You are given a user query you are trying to complete.
You are also given the function to be used and a description of its behavior and input parameter.
Based on this, you should identify which variables (words in the query) to assign to each function parameter in order to solve the query.
You should only return the parameter names and associated variables.
You MUST put the matched parameter names and variables in the format of arg_name1=variable1, arg_name2=variable2, ...
The parameter names MUST be among the ones listed in the Input.
For optional parameter, defined with is_optional=True, you can choose to either assign a variable, e.g.: 'arg_name1=variable1,' or instead leave the parameter unused.
You SHOULD NOT include any other text in the response.
You are given below one example query for the same function, paired with the expected answer to solve the query.
You can use this as an example to predict the correct parameter and variables matching for the query.

(*@\textit{\textcolor{gray}{\# Example Task}}@*)
Input:
User query: 'Open the location "The Peninsula, New York" in Apple Maps and append the summary of the "TravelGuide.pdf" to the note titled "Vacation Plans" in the "Travel" folder.'.
Executed function calls so far with their respective output variable:
out_1 = maps_open_location(query='The Peninsula, New York')
out_2 = open_and_get_file_path(name_or_path='TravelGuide.pdf')
out_3 = summarize_pdf(pdf_path='out_2')
You will be using the function 'append_note_content' to solve this query. What this function does is 'Appends content to an existing note'.
This function accepts the following parameters names, each listed with its expected variable type, whether it is optional, and a short description.
[name: 'name', type: 'str', is_optional: 'False', description: ''], [name: 'append_content', type: 'str', is_optional: 'False', description: ''], [name: 'folder', type: 'str', is_optional: 'True', description: '']
You will pair each of the parameter names above with one variable (or possibly an empty string, if optional). 
Output:

(*@\textit{\textcolor{gray}{\# Example Answer}}@*)
name=Vacation Plans, append_content=out_3, folder=Travel

(*@\textit{\textcolor{gray}{\# Test Task}}@*)
Test query:
Input:
User query: 'Summarize the "Project Proposal.pdf" report, create a reminder to review it by next Tuesday at 2:00 PM, append key points to the note "Project Ideas" in the "Ideas" folder.'.
Executed function calls so far with their respective output variable:
out_1 = open_and_get_file_path(name_or_path='Project Proposal.pdf')
out_2 = summarize_pdf(pdf_path='out_1')
out_3 = create_reminder(name='Review Project Proposal', due_date='2024-02-27 14:00:00', notes='', list_name='', priority=0, all_day=False)
You will be using the function 'append_note_content' to solve this query. What this function does is 'Appends content to an existing note'.
This function accepts the following parameters names, each listed with its expected variable type, whether it is optional, and a short description.
[name: 'name', type: 'str', is_optional: 'False', description: ''], [name: 'append_content', type: 'str', is_optional: 'False', description: ''], [name: 'folder', type: 'str', is_optional: 'True', description: '']
You will pair each of the parameter names above with one variable (or possibly an empty string, if optional).
Output:
\end{promptbox}

\begin{iclbox}{No ICL}
(*@\textit{\textcolor{gray}{\# Test Functions}}@*)
Set of available functions and their descriptions:
[{'function': 'create_calendar_event', 'description': 'Create a calendar event'}, {'function': 'get_phone_number', 'description': 'Search for a contact by name. Returns the phone number of the contact.'}, {'function': 'get_email_address', 'description': 'Search for a contact by name. Returns the email address of the contact'}, {'function': 'compose_new_email', 'description': 'Composes a new email and returns the status of the email composition'}, {'function': 'reply_to_email', 'description': 'Replies to the currently selected email in Mail with the given content'}, {'function': 'forward_email', 'description': 'Forwards the currently selected email in Mail with the given content'}, {'function': 'maps_open_location', 'description': 'Opens the specified location in Apple Maps'}, {'function': 'maps_show_directions', 'description': 'Show directions from a start location to an end location in Apple Maps'}, {'function': 'create_note', 'description': 'Creates a new note with the given content'}, {'function': 'open_note', 'description': 'Opens an existing note by its name'}, {'function': 'append_note_content', 'description': 'Appends content to an existing note'}, {'function': 'create_reminder', 'description': 'Creates a new reminder and returns the status of the reminder creation'}, {'function': 'send_sms', 'description': 'Send an SMS to a list of phone numbers'}, {'function': 'open_and_get_file_path', 'description': 'Opens the file and returns its path'}, {'function': 'get_zoom_meeting_link', 'description': 'Creates a Zoom meeting and returns the join URL'}, {'function': 'summarize_pdf', 'description': 'Summarizes the content of a PDF file and returns the summary'}, {'function': 'join', 'description': 'Collects and combines results from prior actions. It should always be the last action in the plan'}]
You will choose one of the function names above.

(*@\textit{\textcolor{gray}{\# Test Task}}@*)
User query: 'Create a new note titled "Recipe Ideas" in the "Cooking" folder with a list of ingredients for a lasagna recipe. Then, append the steps for preparation to the same note. Finally, send an SMS to yourself with a reminder to buy the ingredients for the lasagna.'.
Executed plan so far:
out_1 = create_note(name='Recipe Ideas', content='Lasagna Ingredients: - Ground beef - Lasagna noodles - Ricotta cheese - Mozzarella cheese - Parmesan cheese - Tomato sauce - Garlic - Onion - Olive oil - Salt - Pepper - Italian seasoning', folder='Cooking')
\end{iclbox}

\begin{iclbox}{Soft Mask}
Set of available functions and their descriptions:
[{'function': 'create_calendar_event', 'description': 'Create a calendar event'}, {'function': 'get_phone_number', 'description': 'Search for a contact by name. Returns the phone number of the contact.'}, {'function': 'get_email_address', 'description': 'Search for a contact by name. Returns the email address of the contact'}, {'function': 'compose_new_email', 'description': 'Composes a new email and returns the status of the email composition'}, {'function': 'reply_to_email', 'description': 'Replies to the currently selected email in Mail with the given content'}, {'function': 'forward_email', 'description': 'Forwards the currently selected email in Mail with the given content'}, {'function': 'maps_open_location', 'description': 'Opens the specified location in Apple Maps'}, {'function': 'maps_show_directions', 'description': 'Show directions from a start location to an end location in Apple Maps'}, {'function': 'create_note', 'description': 'Creates a new note with the given content'}, {'function': 'open_note', 'description': 'Opens an existing note by its name'}, {'function': 'append_note_content', 'description': 'Appends content to an existing note'}, {'function': 'create_reminder', 'description': 'Creates a new reminder and returns the status of the reminder creation'}, {'function': 'send_sms', 'description': 'Send an SMS to a list of phone numbers'}, {'function': 'open_and_get_file_path', 'description': 'Opens the file and returns its path'}, {'function': 'get_zoom_meeting_link', 'description': 'Creates a Zoom meeting and returns the join URL'}, {'function': 'summarize_pdf', 'description': 'Summarizes the content of a PDF file and returns the summary'}, {'function': 'join', 'description': 'Collects and combines results from prior actions. It should always be the last action in the plan'}]
You will choose one of the function names above.

(*@\textit{\textcolor{gray}{\# Soft Mask guidelines}}@*)
To help guide your decision, here is a set of functions extracted from ground truth sequences of training queries.
The functions in this list are functions that have come right after your latest function call history of "('start', 'create_note')" in the training queries. The list is below:
['append_note_content', 'join']
Please use this to guide your decision-making on function selection.

(*@\textit{\textcolor{gray}{\# Test Task}}@*)
User query: 'Create a new note titled "Recipe Ideas" in the "Cooking" folder with a list of ingredients for a lasagna recipe. Then, append the steps for preparation to the same note. Finally, send an SMS to yourself with a reminder to buy the ingredients for the lasagna.'.
Executed plan so far:
out_1 = create_note(name='Recipe Ideas', content='Lasagna Ingredients: - Ground beef - Lasagna noodles - Ricotta cheese - Mozzarella cheese - Parmesan cheese - Tomato sauce - Garlic - Onion - Olive oil - Salt - Pepper - Italian seasoning', folder='Cooking')
\end{iclbox}

\begin{iclbox}{Hard Mask}
(*@\textit{\textcolor{gray}{\# Test Functions after Hard Mask}}@*)
Set of available functions and their descriptions:
[{'function': 'append_note_content', 'description': 'Appends content to an existing note'}, {'function': 'join', 'description': 'Collects and combines results from prior actions. It should always be the last action in the plan'}]
You will choose one of the function names above.

(*@\textit{\textcolor{gray}{\# Test Task}}@*)
User query: 'Create a new note titled "Recipe Ideas" in the "Cooking" folder with a list of ingredients for a lasagna recipe. Then, append the steps for preparation to the same note. Finally, send an SMS to yourself with a reminder to buy the ingredients for the lasagna.'.
Executed plan so far:
out_1 = create_note(name='Recipe Ideas', content='Lasagna Ingredients: - Ground beef - Lasagna noodles - Ricotta cheese - Mozzarella cheese - Parmesan cheese - Tomato sauce - Garlic - Onion - Olive oil - Salt - Pepper - Italian seasoning', folder='Cooking')
\end{iclbox}

\begin{iclbox}{Hard (Weighted)}
(*@\textit{\textcolor{gray}{\# Test Functions after Hard Mask + Weights}}@*)
Set of available functions, their descriptions, and the percentage of times they appear in the ground truth sequences of training queries right after this current function call history:
[{'function': 'append_note_content', 'description': 'Appends content to an existing note', 'probability': 0.714}, {'function': 'join', 'description': 'Collects and combines results from prior actions. It should always be the last action in the plan', 'probability': 0.286}]
You will choose one of the function names above.

(*@\textit{\textcolor{gray}{\# Test Task}}@*)
User query: 'Create a new note titled "Recipe Ideas" in the "Cooking" folder with a list of ingredients for a lasagna recipe. Then, append the steps for preparation to the same note. Finally, send an SMS to yourself with a reminder to buy the ingredients for the lasagna.'.
Executed plan so far:
out_1 = create_note(name='Recipe Ideas', content='Lasagna Ingredients: - Ground beef - Lasagna noodles - Ricotta cheese - Mozzarella cheese - Parmesan cheese - Tomato sauce - Garlic - Onion - Olive oil - Salt - Pepper - Italian seasoning', folder='Cooking')
\end{iclbox}

\section{Use of Large Language Models}
Microsoft Co-Pilot was used for searching for related works and providing edit suggestions to the manuscript.

\onecolumn

% \begin{algorithm}[H]
% \caption{GetFunctionSelectionTrajectory}
% \KwIn{
%   $D$: ground-truth demonstration data; \\
%   $q$: test query; \\
%   $T$: ground-truth test trajectory; \\
%   $M$: agent model; \\
%   $\kwargs$: retrieval/prompting parameters
% }
% \KwOut{$\text{func\_history}$}

% $\text{func\_history} \gets [\,]$; \quad $\text{next\_func} \gets \emptyset$; \quad $i \gets 0$ \;

% \While{$\text{next\_func} \neq \text{end} \ \textbf{or}\ i < \text{MAX\_ITER}$}{
%     $\text{next\_func} \gets \textsc{GetNextFunctionName}(D, q, M,$ \\
%     \hspace{2.5em} $\text{func\_history}=T[0{:}i], \kwargs)$ \;
%     $\text{func\_history}.\text{append}(\text{next\_func})$ \;
%     $i \gets i + 1$ \;
% }

% \Return{$\text{func\_history}$}
% \end{algorithm}

% \begin{algorithm}
% 	\caption{getFunctionSelectionPrompt}
% 	\label{alg:get_fs_prompt}
% 	\begin{algorithmic}[1]
%         \STATE \textbf{Output:} $\text{p}$: function selection prompt
% 		\STATE \textbf{Input:}  F\_retrieved: list of relevant function names; $I$: string specifying ICL method; $W$: boolean specifying to provide probabilities
% 		%\REPEAT
% 		% %$\STATE Initialize $noChange = true$.
%         \IF $I == $ ``no_icl'':
%         \STATE $p \leftarrow$ 
%         \ENDIF
% 		%\UNTIL{$noChange$ is $true$}
% 	\end{algorithmic}
% \end{algorithm}

\section{Pseudocode}\label{app:pseudocode}

While we are currently unable to release the full implementation, we aim to support reproducibility by providing detailed pseudocode that outlines the experimental pipeline, as well as the methods and baselines employed.

\subsection{Overall Experimental Pipeline}

Algorithm \ref{alg:gfst} detail the overall pipeline on how we select multiple function names and subsequently Algorithm \ref{alg:parameter_filling} details how we fill in the parameters for end-to-end function calling.

% \begin{algorithm}
% 	\caption{Get Next Function Name}
% 	\label{alg:get_next_function}
% 	\begin{algorithmic}[1]
%         \STATE \textbf{Output:} $\text{next\_func}$: string of function name selected
% 		\STATE \textbf{Input:}  $D$: ground-truth demonstration data; $q$: test query; func\_history: list of function names so far; $M$: agent model; MAX\_ITERS: number of max function selection iterations; $R$: retrieval method
% 		%\REPEAT
% 		% %$\STATE Initialize $noChange = true$.

%         \STATE \textbf{Return:} next\_func
% 		%\UNTIL{$noChange$ is $true$}
% 	\end{algorithmic}
% \end{algorithm}

\begin{algorithm}
	\caption{Get Function Selection Trajectory}
	\label{alg:gfst}
	\begin{algorithmic}[1]
        \STATE \textbf{Output:} $\text{func\_trajectory}$: list of selected function names for task
		\STATE \textbf{Input:}  query2demonstration: dictionary containing query key with corresponding value being a ground-truth trajectory demonstration; $q$: test query; $T$: ground-truth trajectory of $q$; $M$: agent model
		%\REPEAT
		% %$\STATE Initialize $noChange = true$.
        \STATE Initialize empty list func\_history $\leftarrow []$
        \STATE Initialize empty string new\_func $\leftarrow$ ``''
        \STATE Initialize iter $\leftarrow 0$
        \WHILE{new\_func $\neq$ ``end'' or iter $<$ MAX\_ITERS}
        \STATE func\_history $\leftarrow T[:iter]$
        \STATE Get retrieved functions $\omega$ using either retrieval method; see Sections \ref{app:dtdr_imp} and \ref{app:qts_imp}
        \STATE Construct prompt $p$ using Function Selection Prompt and ICL Method (See Appendix \ref{app:prompt} for more details)
        \STATE Get next\_func $\leftarrow M(p)$
        \STATE $\text{iter} \leftarrow \text{iter} + 1$
        \STATE Append next\_func to func\_trajectory
        \ENDWHILE
        \STATE \textbf{Return:} func\_trajectory
		%\UNTIL{$noChange$ is $true$}
	\end{algorithmic}
\end{algorithm}

\begin{algorithm}
	\caption{Parameter Filling}
	\label{alg:parameter_filling}
	\begin{algorithmic}[1]
        \STATE \textbf{Output:} $\text{func\_param\_trajectory}$: list of dictionaries
		\STATE \textbf{Input:}  $q$: test query; func\_trajectory: list of selected function names so far; func2signatures: dictionary of containing parameter names, types, and constraints for each functin name; $M$: agent model
		%\REPEAT
		% %$\STATE Initialize $noChange = true$.
        \STATE Initialize func\_param\_history $\leftarrow []$
        \STATE Initialize next\_params $\leftarrow \{\}$
        \STATE Initialize iter $\leftarrow 0$
        \WHILE{iter $< \text{len(func\_trajectory)}$}
        \STATE func\_history $\leftarrow T[:\text{iter}]$
        \STATE Set target\_func $\leftarrow$ func\_trajectory[iter] 
        \STATE Set target\_signature $\leftarrow$ func2signatures[target\_func]
        \STATE Get parameter filling prompt $p$ (See Appendix \ref{app:prompt} for more details)
        \STATE Get next\_params = $M(p)$
        \STATE Append to next\_params to func\_param\_trajectory
        \STATE iter $\leftarrow \text{iter} + 1$
        \ENDWHILE
        \STATE \textbf{Return:} func\_param\_trajectory
		%\UNTIL{$noChange$ is $true$}
	\end{algorithmic}
\end{algorithm}

\subsection{DTDR}\label{app:dtdr_imp}

\paragraph{DTDR-C.} Algorithm \ref{alg:build_mc} describes the implementation for building an $N$-order Markov Chain from demonstrations used for DTDR-C, and Algorithm \ref{alg:dtdrc} explains DTDR-C using pre-trained clusters from $k$-Means. \textbf{Static Dependency Retrieval} is implemented by setting the number of clusters $k = 1$ and using order $N = 1$.

\begin{algorithm}
	\caption{Build $N$-order Markov Chain, see Table \ref{tab:tool_dependency_graph_example} for an example}
	\label{alg:build_mc}
	\begin{algorithmic}[1]
        \STATE \textbf{Output:} $G$: nested dictionary of type Dict[Tuple[string] T, Dict[string s, float f]], where T is the previous $N$ function names; s is a function name; float f is a value between $0$ and $1$.
        \STATE \textbf{Input:} $D$ ground-truth demonstration data; $N$: order of the Markov Chain.
		% %$\STATE Initialize $noChange = true$.
        \STATE Initialize $G \leftarrow \{\}$
        \FOR{trajectory in $D$}
        \FOR{$0<=$ index $<L,$ length of trajectory}
        \IF {iter $< N$}
        \STATE Set left-padded $T =$ tuple([`start' ] * ($N-$iter) + trajectory[:index])
        \ELSE
        \STATE Set $T =$ tuple(trajectory[index-(N-1):index])
        \ENDIF
        \IF{$T \notin G$}
        \STATE $G[T] \leftarrow \{\}$
        \ENDIF
        \STATE Set next function $s = $ trajectory[index+1]
        \IF{$s \notin G[T]$}
        \STATE $G[T][s] \leftarrow 0$
        \ENDIF
        \STATE Add occurence to $G$, $G[T][s] \leftarrow G[T][s] + 1$
        \ENDFOR
        \ENDFOR
        \FOR{$T \in G$}
        \STATE Normalize the values in $G[T]$ to sum to $1$
        \ENDFOR
        \STATE \textbf{Return:} G

		%\UNTIL{$noChange$ is $true$}
	\end{algorithmic}
\end{algorithm}

\begin{algorithm}
	\caption{DTDR-Clustering}
	\label{alg:dtdrc}
	\begin{algorithmic}[1]
        \STATE \textbf{Output:} $\omega$: dictionary containing function name key and float value between $0$ and $1$. The sum of all float values should be $1$. 
		\STATE \textbf{Input}: query2demonstration: dictionary containing query key with corresponding value being a ground-truth trajectory demonstration; $q$: test query; func\_history: list of selected function names so far; $N$: order of the Markov Chain; $E$: embedding model; embedding2query: dictionary where keys are embeddings of training queries outputted by $E$ and values are the original input queries; $C$: pre-trained $K$ clusters of embeddings 
		% %$\STATE Initialize $noChange = true$.

        \STATE Set test\_embedding = $E(q)$
        \STATE Find the closest cluster index $0 < k < K$ to test\_embedding
        \STATE Initialize list of demonstrations $D_{k} \leftarrow []$
        \FOR {embedding $\in C_{\text{closest}}$}
        \STATE Get string query $\leftarrow$ embedding2query[embedding]
        \STATE Append query2demonstration$[\text{query}]$ to $D_c$ 
        \ENDFOR
        \STATE Get $N$-order Markov Chain $G$ using Algorithm \ref{alg:build_mc}
        \STATE Set $\omega \leftarrow G[\text{tuple(func\_history)}]$ 
        \STATE \textbf{Return:} $\omega$
		%\UNTIL{$noChange$ is $true$}
	\end{algorithmic}
\end{algorithm}

\paragraph{DTDR-L.} Algorithm \ref{alg:train_f_class} describes the implementation for training the $1$-linear-layer function classifier and Aglorithm \ref{alg:dtdrl} explains the implementation for DTDR-L. \textbf{Static Learned Retrieval} is trained using only the query as input. Therefore, it is done only once at the beginning of task in Algorithm \ref{alg:gfst}.

\begin{algorithm}
	\caption{Train linear layer for function predictino}
	\label{alg:train_f_class}
	\begin{algorithmic}[1]
        \STATE \textbf{Output:} $\phi$: trained $1$-linear-layer classifier
		\STATE \textbf{Input:}  query2demonstration: dictionary containing query key with corresponding value being a ground-truth trajectory demonstration; $E$: embedding model; func\_history: list of selected function names so far
        \FOR{query, demonstration\_trajectoy $\in$ query2demonstration}
        \FOR{$0 <=$ iter $< L,$ length of demonstration\_trajectory}
        \STATE Concatenate $x \leftarrow $ query $+$ demonstration\_trajectory[0:iter-1]
        \STATE Get embedding $e \leftarrow E(x)$
        \STATE Get softmax\_output $\leftarrow Z(e)$
        \STATE Train $\phi$ using Equation \ref{eq:bce_loss}
        \ENDFOR
        \ENDFOR
        \STATE \textbf{Return:} $\phi$
		%\UNTIL{$noChange$ is $true$}
	\end{algorithmic}
\end{algorithm}

\begin{algorithm}
	\caption{DTDR-Linear}
	\label{alg:dtdrl}
	\begin{algorithmic}[1]
        \STATE \textbf{Output:} $\omega$: dictionary containing function name key and float value between $0$ and $1$. The sum of all float values should be $1$. 
		\STATE \textbf{Input:}  $E$: embedding model; $q$: test query; func\_history: list of selected function names so far; $\alpha$: float threshold; $\phi$: trained function name classifier using Algorithm \ref{alg:train_f_class}
        \STATE Concatenate $x \leftarrow q +$ func\_history
        \STATE Get test\_embedding $\leftarrow E(x)$
        \STATE Get softmax\_output $\leftarrow \phi($test\_embedding$)$
        \STATE Get func2softmax, dictionary mapping function name to value from softmax\_output
        \STATE Initialize $\omega$ $\leftarrow \{\}$
        \FOR{func\_name, softmax in func2softmax}
        \IF{softmax $> \alpha$}
        \STATE $\omega$[func\_name] $\leftarrow$ softmax
        \ENDIF
        \ENDFOR
        \STATE Normalize values in $\omega$ to sum up to $1$
        \STATE \textbf{Return:} $\omega$
		%\UNTIL{$noChange$ is $true$}
	\end{algorithmic}
\end{algorithm}

\subsection{Query-Tool Similarity Baselines.}\label{app:qts_imp}
Algorithm \ref{alg:qts} explains the implementation for the Query-Tool Similarity (QTS) retrieval methods. The following methods include vanilla QTS, Less-is-More Level 1, and Tool Graph Retriever. 

\begin{algorithm}
	\caption{Query Tool Similarity Baselines, \textbf{Note: }Unlike DTDR and static DR, the QTS baselines are only done ONCE at the beginning of the task. The retrieved set of functions is constant throughout the function selection steps}
	\label{alg:qts}
	\begin{algorithmic}[1]
        \STATE \textbf{Output:} $\omega$: dictionary containing function name key and float value between $0$ and $1$. The sum of all float values should be $1$. 
		\STATE \textbf{Input:}  func\_descriptions: dictionary containing function name key and the corresponding documentation; $E$: embedding model; $q$: test query; qts\_method: string specifying which QTS baseline to use;
        $M$: agent model to generate what the agent thinks are the relevant tools for $q$ if qts\_method $==$ ``less\_is\_more''; $G_T$ : matrix representing a tool dependency graph. We obtain $G_T$ via Algorithm \ref{alg:build_mc} by setting $N = 1$ and $K=1$;
        $\alpha$: float threshold
		%\REPEAT
		% %$\STATE Initialize $noChange = true$.
        \IF{qts\_method == ``less\_is\_more''}
        \STATE Set Less is More Thought Prompt $p$
        \STATE Set string\_to\_embed $\leftarrow M(q, p)$
        \ELSE
        \STATE string\_to\_embed $\leftarrow q$
        \ENDIF
        \STATE test\_embedding $\leftarrow E(\text{string\_to\_embed})$

        \STATE Set func\_description\_embedding $\leftarrow \{\}$
        \FOR{func\_name, description $\in$ func\_description}
        \STATE func\_description\_embeddings[func\_name] $\leftarrow$ $E($description$)$
        \ENDFOR

        \IF{qts\_method $==$ ``tool\_graph\_retrieval''}
        \STATE  Update func\_description\_embeddings with $G_T$ using convolution operation
        \ENDIF

        \STATE Initialize query\_tool\_distances $\leftarrow \{\}$
        \FOR{func\_name, func\_embedding $\in$ func\_description\_embeddings}
        \STATE query\_tool\_distances[func\_name] $\leftarrow$ cosine\_similarity(test\_embedding, func\_embedding) 
        \ENDFOR
        \STATE Initialize $\omega$ $\leftarrow \{\}$
        \FOR{func\_name, similarity in query\_tool\_distances}
        \IF{similarity $> \alpha$}
        \STATE $\omega$[func\_name] $\leftarrow$ similarity
        \ENDIF
        \ENDFOR
        \STATE Normalize values in $\omega$ to sum up to $1$
        \STATE \textbf{Return:} $\omega$
		%\UNTIL{$noChange$ is $true$}
	\end{algorithmic}
\end{algorithm}

% def get_parameter_filling_values(test_query, predicted_function_sequence, agent_model, function_signatures,**kwargs) -> list[dict[string:string]]:
    
%     Initialize param_history = []
%     Initialize next_params = {}
%     Intialize iter = 0
%     While iter < len(predicted_function_sequence):
%         # Identify current target function and its signature
%         Set target_function = predicted_function_sequence[iter]
%         Set target_signature = function_signatures[target_func]
        
%         # This function creates a prompt string based on the template as shown in Appendix G “Full Prompt G.2: Parameter Filling.”

%         Get parameter_filling_prompt = format_prompt(
%             test_query = test_query,
%             target_function = target_function,
%             target_signature = target_signature,
%             func_history = predicted_function_sequence[:iter],     # function names up to current step
%             param_history = param_history[:iter],                 # ground-truth params up to current step
%             **kwargs 			# for instance, ICL examples can be included in the prompt
%         )

%         # Request parameters for the current function using the agent model
%         Get next_params = agent_model(parameter_filling_prompt)

%         # Append and advance
%         param_history.append(next_params)
%         Increment iter += 1

\end{document}